\documentclass{article}

\usepackage[
            CJKbookmarks=true,
            bookmarksnumbered=true,
            bookmarksopen=true,
            colorlinks=true,
                        citecolor=magenta,
            linkcolor=blue,
            anchorcolor=red,
            urlcolor=blue
            ]{hyperref}

\usepackage[utf8]{inputenc} 
\usepackage[T1]{fontenc}    
\usepackage{url}            
\usepackage{booktabs}       
\usepackage{amsfonts}       
\usepackage{nicefrac}       
\usepackage{microtype}      

\usepackage{graphicx,color,fancyhdr}
\usepackage{amsmath,amssymb,amsthm,amsfonts,dsfont}
\usepackage{subfig}
\usepackage{enumerate}
\usepackage{algorithmic, algorithm}
\usepackage{mathtools}

\newcommand{\Expect}{\mathbb{E}}
\newcommand{\ud}{\,\mathrm{d}}

\newcommand{\norm}[1]{\|#1\|}

\renewcommand{\tilde}{\widetilde}
\newcommand{\cvx}{\mathtt{CVX}}
\newcommand{\icvx}{\mathtt{ICNN}}

\newcommand{\real}{\mathbb{R}}

\newcommand{\calP}{\mathcal{P}}

\newcommand{\calX}{\mathcal{X}}
\newcommand{\calY}{\mathcal{Y}}

\usepackage{thmtools}

\declaretheorem[numberwithin=section]{theorem}

\declaretheorem[sibling=theorem]{remark}

\usepackage{prettyref,xspace,graphicx}
\usepackage{bbm,tikz}

\newrefformat{cond}{Condition~\ref{#1}}
\newrefformat{eq}{(\ref{#1})}
\newrefformat{thm}{Theorem~\ref{#1}}
\newrefformat{th}{Theorem~\ref{#1}}
\newrefformat{chap}{Chapter~\ref{#1}}
\newrefformat{sec}{Section~\ref{#1}}
\newrefformat{algo}{Algorithm~\ref{#1}}
\newrefformat{fig}{Figure~\ref{#1}}
\newrefformat{tab}{Table~\ref{#1}}
\newrefformat{rmk}{Remark~\ref{#1}}
\newrefformat{clm}{Claim~\ref{#1}}
\newrefformat{def}{Definition~\ref{#1}}
\newrefformat{cor}{Corollary~\ref{#1}}
\newrefformat{lmm}{Lemma~\ref{#1}}
\newrefformat{prop}{Proposition~\ref{#1}}
\newrefformat{pr}{Proposition~\ref{#1}}
\newrefformat{app}{Appendix~\ref{#1}}
\newrefformat{prob}{Problem~\ref{#1}}
\newrefformat{ques}{Question~\ref{#1}}
\newrefformat{note}{Note~\ref{#1}}
\newrefformat{assump}{Assumption~\ref{#1}}
\newrefformat{issue}{Issue ~\ref{#1}}
\newrefformat{fix}{Fix ~\ref{#1}}

\newcommand{\reals}{\mathbb{R}}

\newcommand{\inner}[2]{\langle {#1}, {#2}  \rangle}

\newcommand{\calB}{\mathcal{B}}

\newcommand{\calN}{\mathcal{N}}

\newcommand{\calV}{\mathcal{V}}

\newcommand{\define}{\triangleq}

\newcommand{\qth}[1]{\left[ #1 \right]}

\newcommand{\ie}{i.e.\xspace}

\newcommand{\Id}{\mathrm{Id}}
\mathchardef\mhyphen="2D

\usepackage[accepted]{icml2020}

\icmltitlerunning{Optimal transport mapping via input convex neural networks}

\begin{document}

\twocolumn[
\icmltitle{Optimal transport mapping via input convex neural networks}



\icmlsetsymbol{equal}{*}

\begin{icmlauthorlist}
\icmlauthor{Ashok Vardhan Makkuva}{equal,uiuc}
\icmlauthor{Amirhossein Taghvaei}{equal,uc}
\icmlauthor{Jason D. Lee}{prince}
\icmlauthor{Sewoong Oh}{uw}
\end{icmlauthorlist}

\icmlaffiliation{uiuc}{Department of Electrical and Computer Engineering, University of Illinois at Urbana-Champaign.}
\icmlaffiliation{uc}{Department of Mechanical and Aerospace Engineering, University of California, Irvine.}
\icmlaffiliation{prince}{Department of Electrical Engineering, Princeton University,}
\icmlaffiliation{uw}{Allen School of Computer Science \& Engineering, University of Washington}

\icmlcorrespondingauthor{Ashok}{makkuva2@illinois.edu}
\icmlcorrespondingauthor{Amir}{amirhoseintghv@gmail.com}

\icmlkeywords{Machine Learning, ICML}

\vskip 0.3in
]



\printAffiliationsAndNotice{\icmlEqualContribution} 

\begin{abstract}
In this paper, we present a novel and principled approach to learn the optimal transport between two distributions, from samples. Guided by the optimal transport theory, we learn the optimal Kantorovich potential which induces the optimal transport map. This involves learning two convex functions, by solving a novel minimax optimization. Building upon recent advances in the field of input convex neural networks, we propose a new framework to estimate the optimal transport mapping as the gradient of a convex function that is trained via minimax optimization. Numerical experiments confirm the accuracy of the learned transport map. Our approach can be readily used to train a deep generative model. When trained between a simple distribution in the latent space and a target distribution, the learned optimal transport map acts as a deep generative model. Although scaling this to a large dataset is challenging, we demonstrate two important strengths over standard adversarial training: robustness and discontinuity. As we seek the optimal transport, the learned generative model provides the same mapping regardless of how we initialize the neural networks. Further, a gradient of a neural network can easily represent discontinuous mappings, unlike standard neural networks that are constrained to be continuous. This allows the learned transport map to match any target distribution with many discontinuous supports and achieve sharp boundaries.
\end{abstract}

\section{Introduction}
\label{sec:introduction}

Finding a mapping that transports mass from one distribution $Q$ to another distribution $P$ 
is an important task in various machine learning applications, such as deep generative models \cite{goodfellow2014generative,2013auto} and domain adaptation \cite{gopalan2011domain,ben2010theory}. 
Among infinitely many transport maps $T$ that can map a random variable $X$ from $Q$ such that 
$T(X)$ is distributed as $P$, several recent advances focus on  
discovering some inductive bias to find a transport map with desirable properties.  
Research in optimal transport has been leading such efforts, 
in applications such as 
color transfer \cite{ferradans2014regularized}, 
shape matching \cite{su2015optimal}, 
 data assimilation \cite{reich2013nonparametric}, 
and  Bayesian inference \cite{el2012bayesian}. 
Searching for an optimal transport 
encourages a mapping  that minimizes the total cost 
of transporting mass from $Q$ to $P$, as originally formulated in \citet{monge1781memoire}, 
and provides the inductive bias needed in many such applications.  
However, finding the optimal transport map in general is a challenging task, especially in high dimensions where 
efficient approaches are critical. 

Algorithmic solutions are well-established 
for discrete variables; 
the optimal transport can be found as a solution to linear program. 
Building upon this mature area,  typical approaches for general distributions use quantization,  
and this becomes intractable for 
high-dimensional variables we encounter in modern applications \cite{evans1999differential,benamou2000computational,papadakis2014optimal}. 
 
 \begin{figure*}[t]
 \vspace{-0.2cm}
 \captionsetup[subfloat]{captionskip=-1.0pt}
\centering
\begin{tabular}{cccc}
	\hspace{-.6cm}
\subfloat[Data samples]{
	\includegraphics[width=0.28\hsize]{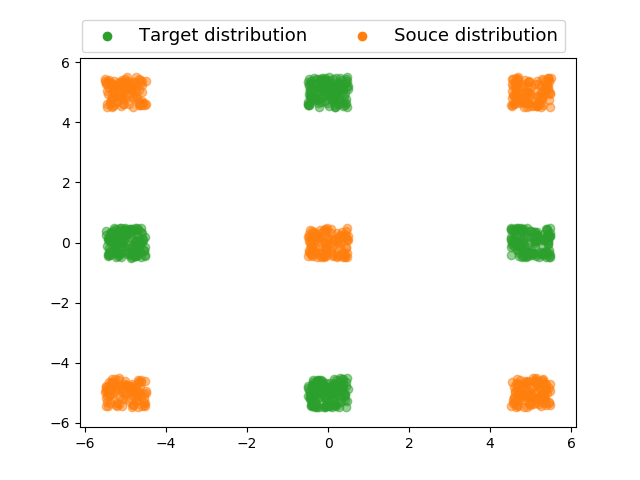}\label{fig:checker-board-data}
	} 
	& 	
	\hspace{-0.9cm}
	\subfloat[Our transport map]{
	\includegraphics[width=0.28\hsize]{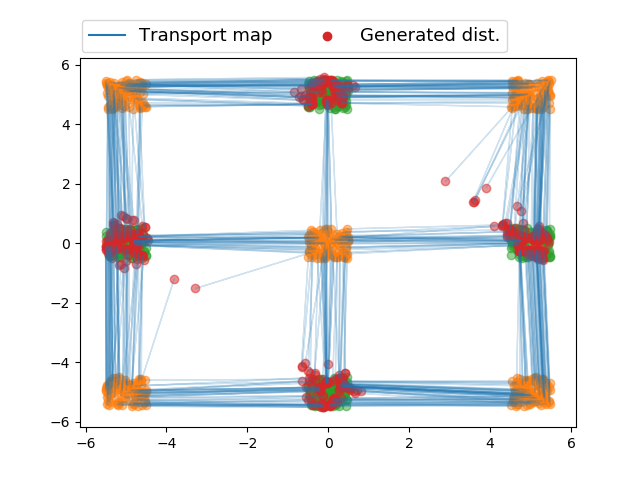}\label{fig:checker-board-OT}
	} 
	&
	\hspace{-1.0cm}
	\subfloat[Displacement vector field]{
		\includegraphics[width=0.28\hsize]{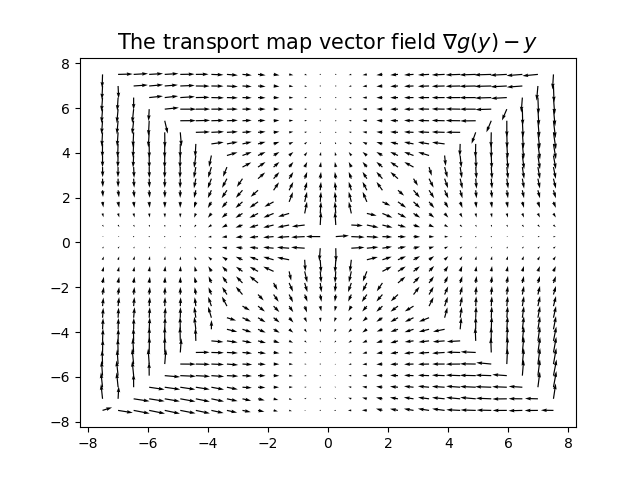}\label{fig:g-vector-field}
		}
		 & 	
		\hspace{-1.0cm}
	\subfloat[Level sets]{
		\includegraphics[width=0.28\hsize]{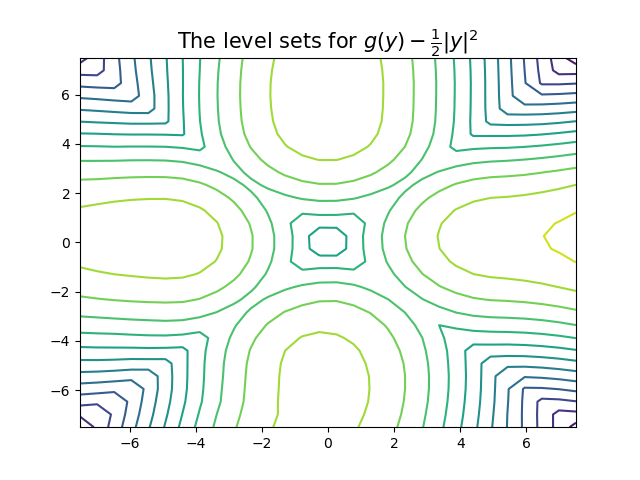}\label{fig:g-level-sets}
		} 
\end{tabular}
\vspace{-0.2cm}
\caption{Results on Checkerboard dataset. (a) Samples from the source (orange) and target (green) distributions; 
(b) The learned transport map and the generated distribution, via Algorithm~\ref{alg:W2}; 
(c) The learned displacement vector field generated by $\nabla g(y)-y$; 
(d) The level sets of the original dual variable $g(y) - \frac{1}{2}|y|^2$. The experimental details are included in Section~\ref{sec:exp_map}.
}
\label{fig:checker-board-OT0}
\vspace{-0.4cm}
\end{figure*}

To this end, we propose a novel minimax optimization approach to search for the 
optimal transport under the quadratic distance (i.e.~2-Wassertstein metric). 
A major challenge in a minimax formulation of optimal transport is 
that the constraints in the Kantorovich dual formulation \eqref{eq:dual_form} are notoriously challenging. 
They require the evaluation of the functions at every point in the domain, which is not tractable. 
A common straightforward heuristics sample some points and add those sampled constraints as regularizers. 
Such regularizations create biases that hinder learning the true optimal transport. 

Our key innovation is to depart from this common practice;   
we instead eliminate the constraints by restricting our search to the set of all convex functions, 
building upon the fundamental connection from Theorem \ref{thm:knott-brenier}. 
This leads to a novel minimax formulation  in \eqref{eq:max-min}. 
Leveraging on recent advances in input convex neural networks, 
we propose a new architecture and a training algorithm for solving this minimax optimization. 
We establish the consistency of our proposed minimax formulation in Theorem~\ref{thm:our_optim_result}. 
In particular, we show that the solution to this optimization problems yields the exact optimal transport map. 
We provide stability analysis for the proposed estimator in \prettyref{thm:stability}.

Further, when used to train deep generative models, 
our approach can be viewed as a novel framework to train 
a generator that is modeled as a {\em gradient of a convex function}. 
We provide a principled training rule based on the optimal transport theory. 
This ensures that $(i)$ the generator converges to the optimal transport, independent of how we initialize the neural network; and 
$(ii)$  represent sharp boundaries when the target has multiple disconnected supports. 
Gradient of a neural network naturally represents discontinuous functions, 
which is critical in mapping from a single connected support to  disconnected supports.

To model convex functions, we leverage Input Convex Neural Networks (ICNNs),  
a class of scalar-valued neural networks $f(x;\theta)$ such that the function $x \mapsto f(x;\theta) \in \reals$ is convex. 
These neural networks were introduced by \citet{amos2016input} to provide efficient inference and optimization procedures for structured prediction, data imputation and reinforcement learning tasks. In this paper, we show that ICNNs can be efficiently trained to learn the optimal transport map between two distributions $P$ and $Q$. To the best of our knowledge, this is the first such instance where ICNNs are leveraged for the well-known task of learning optimal transport maps in a {\em scalable} fashion. This framework opens up a new realm for understanding problems in optimal transport theory using parametric convex neural networks, both in theory and practice. 
Figure~\ref{fig:checker-board-OT0} provides an example 
where the optimal transport map has been learned via our proposed Algorithm~\ref{alg:W2} 
from the orange distribution to the green distribution.

\textbf{Notation.} $\calP(\calX)$ denotes the set of probability measures on a Polish space $\calX$, and $\calB(\calX)$ denotes the Borel subsets of $\calX$.  For $P \in \calP(\calX)$ and $Q\in \calP(\calY)$, $P \otimes Q $ denotes the product measure on $\calX \times \calY$.  For measurable map $T:\calX \to \calY$, $T_{\#} P$ denotes the push-forward of  $P$ under $T$, \ie $(T_{\#} P)(A)=P(T^{-1}(A)),~\forall A \in \calB(\calY)$. 
 $L^1(P) \triangleq \{f \text{ is measurable } \&~ \int f\ud P<\infty\}$ denotes the set of integrable functions with respect to $P$.  $\cvx(P)$ denotes the set of all convex functions in $L^1(P)$. $\mathrm{Id}:x\mapsto x$ denotes the identity function. $\langle \cdot,\cdot \rangle$  and $\|\cdot\|$ denote the inner-product and $\ell_2$-Euclidean norm. 


\vspace{-0.2cm}
\section{Background on optimal transport} 
\label{sec:background} 

 Let $P$ and $Q$ be two probability distributions on $\real^d$ with finite second order moments. 
The \emph{Monge's optimal transportation problem} is to transport the probability mass under $Q$ to $P$ with the least amount of cost\footnote{In general, Monge's problem is defined in terms of cost function $c(x,y)$. This paper is concerned with quadratic cost function $c(x,y)=\frac{1}{2}\|x-y\|^2$ because of its nice geometrical properties and connection to convex analysis~\citep[Ch. 2]{villani2003topics}.}, \ie
\begin{align}
	\underset{T: T_{\#}Q=P} {\text{minimize}}\;\; \;\; \frac{1}{2} \Expect_{X\sim Q}  \norm{X-T(X)}^2. \;
	\label{eq:monge}
\end{align}
Any transport map $T$ achieving the minimum in \prettyref{eq:monge} is called  \emph{optimal transport map}. 
Optimal transport map may not  exist. In fact, the feasible set in the above optimization problem may itself be empty, for example when $Q$ is a Dirac distribution and $P$ is any non-Dirac distribution.

To resolve the existence issue of the Monge problem~\eqref{eq:monge}, 
 Kantorovich introduced a relaxation of the problem,
\begin{align}
W_2^2(P,Q) \triangleq \inf_{\pi \in \Pi(P,Q)}~ \frac{1}{2} \Expect_{(X,Y) \sim \pi}\norm{X-Y}^2,
\label{eq:kantor_relax}
\end{align}
where $\Pi(P,Q)$ denotes the set of all joint probability distributions (or equivalently, couplings) whose first and second marginals are $P$ and $Q$, respectively. The optimal value in~\eqref{eq:kantor_relax} is the $2$-Wasserstein distance $W_2(\cdot,\cdot)$ squared. Any coupling $\pi$ achieving the infimum 
is called the \emph{optimal coupling}. Optimization problem   
\eqref{eq:kantor_relax} is also referred to as the {\em  primal formulation} for $2$-Wasserstein distance.

Kantorovich also provided a dual formulation for \eqref{eq:kantor_relax}, known as the Kantorovich duality \citep[Theorem 1.3]{villani2003topics}, 
\begin{align}
	W_2^2(P,Q) =\; \sup_{(f,g)\in \Phi_c} \Expect_P[f(X)]+\Expect_Q[g(Y)],
	\label{eq:dual_form}
\end{align}
where $\Phi_c$ denotes the constrained space of functions, defined as $\Phi_c \define \bigl\{(f,g)\in L^1(P)\times L^1(Q): ~f(x)+g(y)\leq \frac{1}{2}\norm{x-y}^2_2, \quad \forall (x,y) ~d P \otimes d Q~\text{a.e.}\bigr\}$. 

The dual problem~\eqref{eq:dual_form} can be recast as an stochastic optimization problem by approximating the expectations using independent samples from $P$ and $Q$. However, there is no easy way to ensure the feasibility of the constraint~$(f,g)\in \Phi_c$ along the gradient updates.  
Common approach is to translate the optimization into a tractable form, while sacrificing the original goal of finding the optimal transport map. 
Concretely, an entropic or a quadratic regularizer is added to the primal problem~\eqref{eq:kantor_relax}~\cite{cuturi2013sinkhorn,essid2018quadratically,peyre2019computational,blondel2017smooth}. 
Then, the dual to the regularized primal problem is an unconstrained version of \eqref{eq:dual_form} 
with additional penalty term.
The unconstrained problem  can be numerically solved using Sinkhorn algorithm in discrete setting~\cite{cuturi2013sinkhorn} 
or stochastic gradient methods with suitable function representation in continuous setting~\cite{genevay2016stochastic,seguy2017large}.   
The optimal transport can then be obtained from $f$ and $g$,  using the first-order optimality conditions of the Fenchel-Rockafellar's duality theorem~\cite{seguy2017large}, or by training a generator through an adversarial computational procedure~\cite{leygonie2019adversarial}. 

In this paper, we take a different approach: solve the dual problem without introducing a regularization. This builds upon~\cite{taghvaei20192}, where ICNN for the task of approximating the Wasserstein distance and optimal transport map is originally proposed. We bring the idea proposed~\cite{taghvaei20192} into practice by introducing a novel minimax optimization formulation. We describe our proposed method in~Section~\ref{sec:formulation} and provide a detailed comparison in Remark~\ref{rem:compare}. Discussion about other related works \citep{lei2017geometric,guo2019mode, xie2019scalable, muzellec2019subspace,rabin2011wasserstein, korotin2019wasserstein} appears in \prettyref{app:related-work-more}.

\vspace{-0.3cm}
\section{A novel minimax formulation to learn optimal transport}
\label{sec:formulation}

Our goal is to learn the optimal transport map $T^*$ from $Q$ to $P$, 
from samples drawn from $P$ and $Q$, respectively. 
We use the 
fundamental connection between optimal transport and Kantorovich dual in Theorem~\ref{thm:knott-brenier}, 
to formulate learning $T^*$ as a problem of estimating $W_2^2(P,Q)$. 
However, $W_2^2(P,Q)$ is notoriously hard to estimate. 
The standard Kantorovich dual formulation in Eq.~\eqref{eq:dual_form} involves a 
supremum over a  set $\Phi_c$ with infinite constraints,  
which is challenging to even approximately project onto. 
To this end, we derive an alternative optimization formulation in Eq.~\eqref{eq:max-min}, 
inspired by the convexification trick \citep[Section 2.1.2]{villani2003topics}.
This allows us to eliminate the distance constraint of $\Phi_c$, 
and instead constrain our search over all {\em convex functions}. 
This constrained optimization can now be seamlessly integrated with recent advances in 
designing deep neural architectures with convexity guarantees. 
This leads to a novel minimax optimization 
to learn the optimal transport. 

We exploit the fundamental properties of $W_2^2(P,Q)$ and the corresponding optimal transport to 
reparametrize the optimization formulation. 
Note that for any $(f,g) \in \Phi_c$, 
\begin{align*}
&f(x)+g(y) \leq \frac{1}{2}\norm{x-y}_2^2 \;\; \Longleftrightarrow\\ &\qth{\frac{1}{2}\|x\|_2^2 -f(x)}+\qth{\frac{1}{2}\|y\|_2^2 -g(y)} \geq \inner{x}{y}.
\end{align*}
Hence reparametrizing $\frac{1}{2}\|\cdot \|_2^2-f(\cdot)$ and $\frac{1}{2}\|\cdot\|_2^2-g(\cdot)$ by $f$ and $g$ respectively, and 
substituting them in \prettyref{eq:dual_form}  yields  
\begin{align*}
W_2^2(P,Q)  = C_{P,Q}  -\inf_{(f,g) \in \tilde{\Phi}_c} \Big\{ \Expect_P[f(X)]+\Expect_Q[g(Y)] \Big\},
\end{align*}
where $C_{P,Q}=(1/2)\Expect [\norm{X}_2^2+\norm{Y}_2^2 ]$ is a constant independent of $(f,g)$ and $\tilde{\Phi}_c \define \{ (f,g) \in L^1(P) \times L^1(Q): f(x)+g(y) \geq \inner{x}{y}, \quad  \forall (x,y) ~ dP \otimes dQ ~\text{a.e.} \}$. While the above constrained optimization problem involves a pair of functions $(f,g)$, it can be transformed into the following form involving only a single convex function $f$, thanks to \citet[Theorem 2.9]{villani2003topics}:
\begin{align}\label{eq:dual_convex_form}
\hspace{-2pt}W_2^2(P,Q) \!= \!C_{P,Q}\!-  \!\!\inf_{f \in \cvx(P)} \Expect_P[f(X)]\!+\!\Expect_Q[f^\ast(Y)],
\end{align}
where $f^\ast(y)=\sup_x \langle x,y\rangle - f(x)$ is the convex conjugate of $f(\cdot)$. 

The crucial tools behind our formulation are 
the following celebrated results due to Knott-Smith and Brenier \cite{villani2003topics}, 
which relate the optimal solutions for the dual form  in \prettyref{eq:dual_convex_form} and 
the primal form in \prettyref{eq:kantor_relax}.
\begin{theorem}[{\citep[Theorem 2.12]{villani2003topics}}]
\label{thm:knott-brenier}
Let $P,Q$ be  two probability distributions on $\reals^d$ with finite second order moments. Then,
\begin{enumerate}
\item 
(\textbf{Knott-Smith optimality criterion}) A coupling $\pi \in \Pi(P,Q)$ is optimal for the primal \prettyref{eq:kantor_relax} if and only if there exists a convex function $f \in \cvx(\reals^d)$ such that $\mathrm{Supp}(\pi) \subset \mathrm{Graph}(\partial f)$. Or equivalently, for all $d\pi$-almost $(x,y)$, $y \in \partial f(x)$. Moreover, the pair $(f,f^\ast)$ achieves the minimum in the dual form \prettyref{eq:dual_convex_form}.
\item
 (\textbf{Brenier's theorem}) If $Q$ admits a density with respect to the Lebesgue measure on $\reals^d$, then there is a unique optimal coupling $\pi$ for the primal problem. In particular, the optimal coupling satisfies that
\begin{align*}
d\pi(x,y) =  dQ(y) \delta_{x=\nabla f^\ast(y)},
\end{align*}
where the convex pair $(f,f^\ast) $ achieves the minimum in the dual problem \prettyref{eq:dual_convex_form}. Equivalently, $\pi=(\nabla f^\ast \times \Id)_{\#}Q$.
\item 
Under the above assumptions of Brenier's theorem, $\nabla f^\ast$ in the unique solution to Monge transportation problem from $Q$ to $P$, \ie
\begin{align*}
\Expect_Q \norm{\nabla f^\ast(Y)-Y}^2 = \inf_{T: T_{\#}Q=P}\Expect_Q \norm{T(Y)-Y}^2.
\end{align*}
\end{enumerate}
\end{theorem}
\begin{remark}\normalfont
Whenever $Q$ admits a density, we refer to $\nabla f^\ast$ as the optimal transport map.
\label{rem:main}
\end{remark}

Henceforth, throughout the paper we assume that the distribution $Q$ admits a density in $\reals^d$.
Note that in view of \prettyref{thm:knott-brenier}, any optimal pair $(f,f^\ast)$ from the dual formulation in \prettyref{eq:dual_convex_form} provides us an optimal transport map $\nabla f^\ast$ pushing forward $Q$ onto $P$. However, the objective \prettyref{eq:dual_convex_form} is not amenable to standard stochastic optimization schemes due to the conjugate function $f^\ast$. 
To this end, we propose a novel minimax formulation in the following theorem where we replace the conjugate with a new convex function.
\begin{theorem}
\label{thm:our_optim_result}
Whenever $Q$ admits a density in $\reals^d$, we have 
\begin{align}\label{eq:max-min}
&W_2^2(P,Q) =  \sup_{\substack{f \in \cvx(P), \\ f^\ast \in L^1(Q)}} \inf_{g \in \cvx(Q) }~\calV_{P,Q}(f,g) + C_{P,Q}, 
\end{align}
where 
$\calV_{P,Q}(f,g)$ is a functional of $f,g$ defined as 
\begin{equation*}
\calV_{P,Q}(f,g)= -\Expect_P[f(X)]-\Expect_Q[\inner{Y}{\nabla g(Y)}-f(\nabla g(Y))].
\end{equation*}
In addition, there exists an optimal pair $(f_0, g_0)$ achieving the infimum and supremum respectively, where  $\nabla g_0$ is the optimal transport map from $Q$ to $P$.
\end{theorem}
\vspace{-10pt}
\begin{proof}[Proof sketch]
	The proof follows from the inequality $\langle y, \nabla g(y) \rangle  - f(\nabla g(y)) \leq f^*(y)$ for all functions $g$, and then taking the expectation over $Q$, and observing that the equality is achieved with $g=f^*$. The technical details appear in \prettyref{app:proof_theorem}.
\end{proof}
\begin{remark}\normalfont
\label{rem:relax}
	For any convex function $f$,  the function $g\in L^1(Q)$ that achieves the infimum in~\eqref{eq:max-min} is convex and equals $f^*$. Therefore, the constraint $g\in \cvx(Q)$ can be relaxed to $g\in L^1(Q)$ without changing the optimal value and optimizing functions. We numerically observe that the optimization algorithm performs better under this relaxation.
\end{remark}

Formulation~\eqref{eq:max-min} now provides a principled approach to learn the optimal transport mapping $\nabla g(\cdot)$
as a solution of a minimax optimization. Since the optimization  involves the search over the space of convex functions, we utilize the recent advances in input convex neural networks (ICNNs) to parametrize them as discussed in the following section.

\subsection{Minimax optimization over ICNNs} 
\label{sec:icnn} 

We propose using parametric models based on deep neural networks to approximate the set of convex functions. 
This is known as  input convex neural networks \cite{amos2016input}, denoted by $\icvx(\reals^d)$. 
We propose estimating 
the following approximate Wasserstein-$2$ distance, from samples: 
\begin{align}
\tilde{W}_2^2(P,Q) \!= \!\sup_{f\in \icvx(\real^d)} \inf_{g \in \icvx(\real^d)}\calV_{P,Q}(f,g)\! +\! C_{P,Q}.
\label{eq:approxW2}
\end{align}
ICNNs are a class of scalar-valued neural networks $f(x;\theta)$ such that the function $x \mapsto f(x;\theta) \in \reals$ is convex. 

\begin{figure}[t]
	\centering
	\includegraphics[width=0.95\hsize]{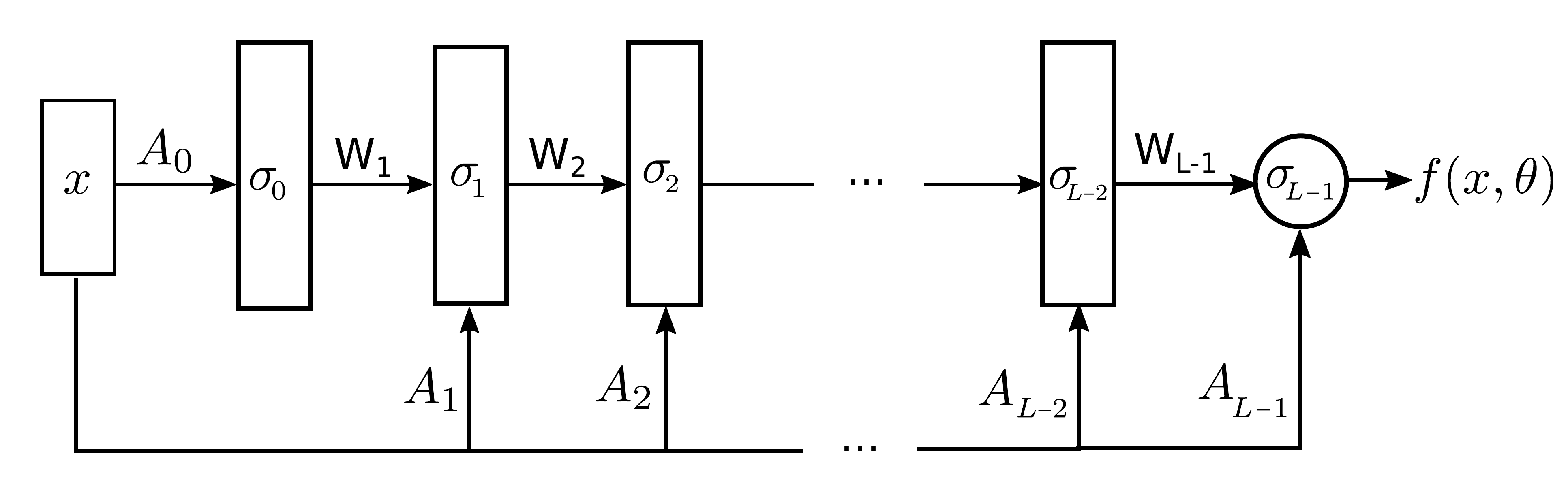}
	\caption{The input convex neural network (ICNN) architecture.}
	\label{fig:ICNN}
	\vspace{-0.3cm}
\end{figure}

The neural network architecture for an ICNN is as follows.
 Given an input $x \in \reals^d$, the mapping $x \mapsto f(x;\theta)$ is given by a $L$-layer feed-forward NN using the following equations for $l=0,1,\ldots, L-1$:
\begin{align*}
z_{l+1} = \sigma_l(W_l z_l + A_l x + b_l),\quad f(x;\theta)=z_L,
\end{align*}
where $\{W_l\}$, $\{A_l\}$ are weight matrices (with the convention that $W_0=0$), and $\{b_l\}$ are the bias terms. $\sigma_l$ denotes the entry-wise activation function at the layer $l$. 
This is illustrated in Figure~\ref{fig:ICNN}. 
We denote the total set of parameters by $\theta=(\{W_l\},\{A_l\},\{b_l\})$. It follows from \citet[Proposition 1]{amos2016input} that $f(x;\theta)$ is convex in $x$ provided

(i) all entries of the weights $W_l$ are non-negative;

(ii) activation function $\sigma_0$ is convex; 

(iii) $\sigma_l$ is convex and non-decreasing, for $l=1,\ldots,L-1$.

While ICNNs are a specific parametric class of convex functions, it is important to understand if this class is rich enough representationally. This is answered positively by \citet[Theorem 1]{chen2018optimal}. In particular, they show that any convex function over a compact domain can be approximated in sup norm by a ICNN to the desired accuracy. This justifies the choice of ICNNs as a suitable approximating class for the convex functions.

\begin{figure*}[t]

	\centering
	\begin{tabular}{cccc}
		\hspace{-0.5cm}
			\includegraphics[trim=55 0 45 0, clip,width=0.24\hsize]{{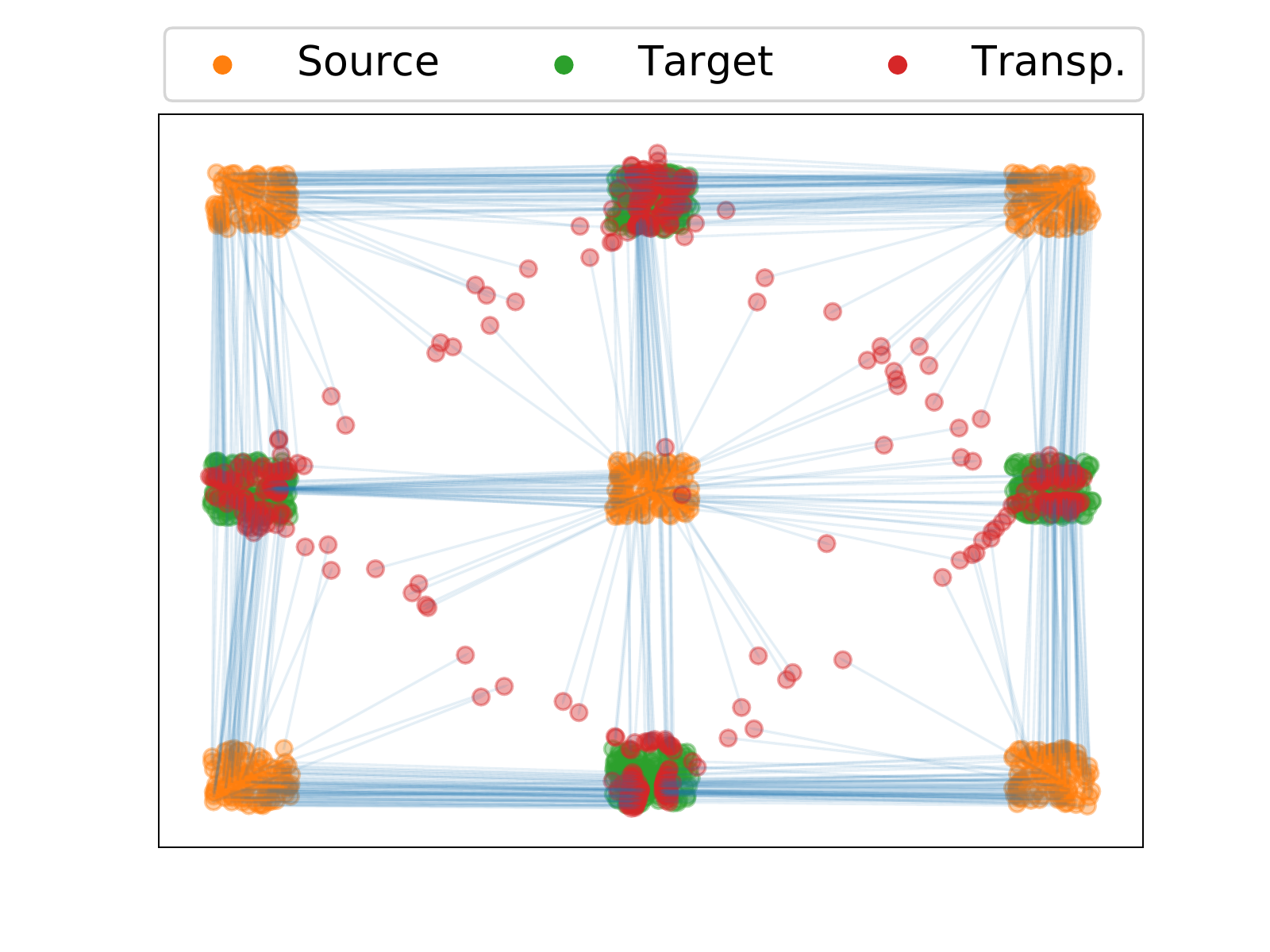}}
			\label{fig:8gaussians-data}
		& 	
		\hspace{-0.35cm}
			\includegraphics[trim=55 0 45 0, clip,width=0.24\hsize]{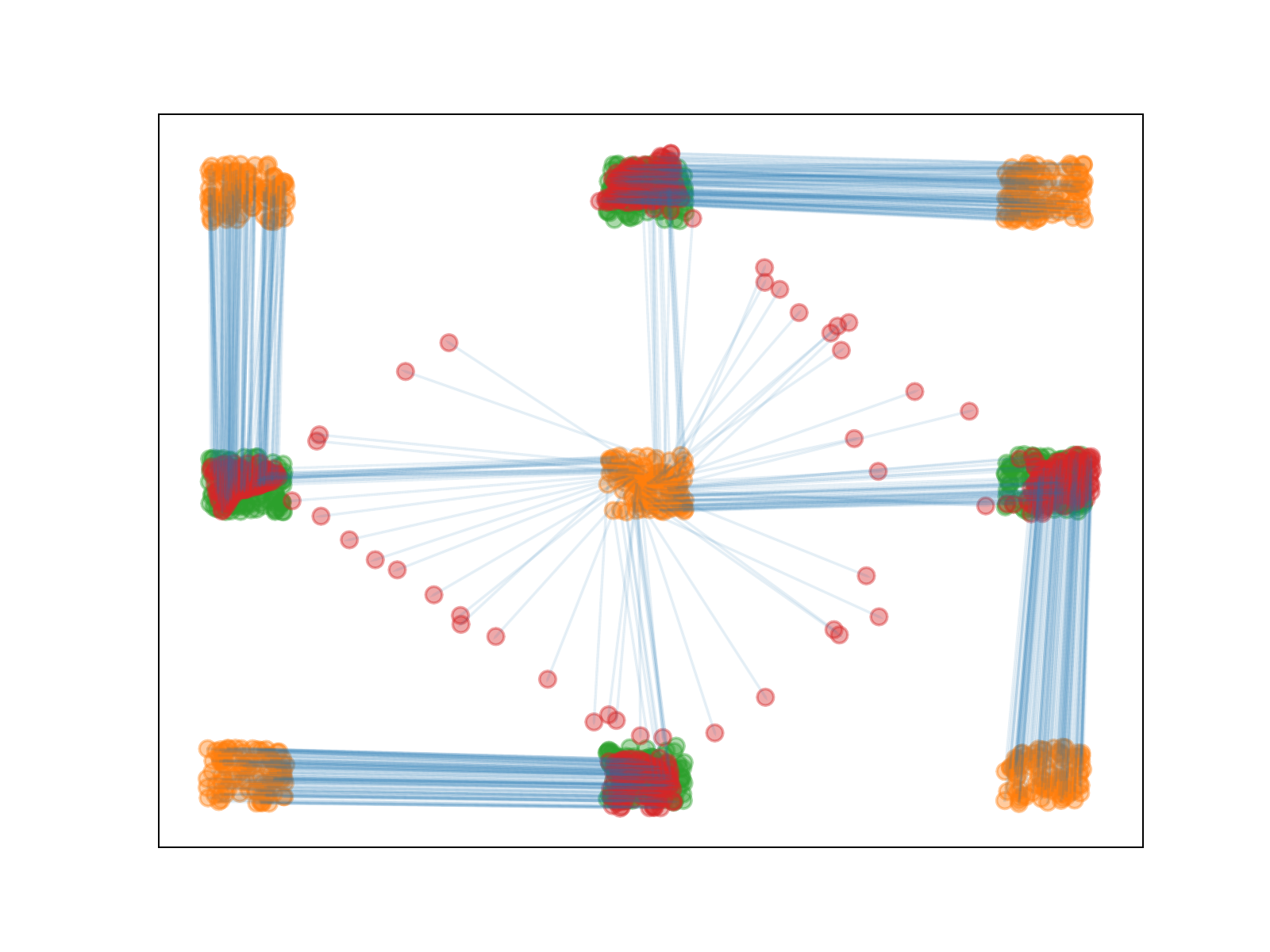}
			\label{fig:circle-data}
			& 	
		\hspace{-0.35cm}
			\includegraphics[trim=55 0 45 0, clip,width=0.24\hsize]{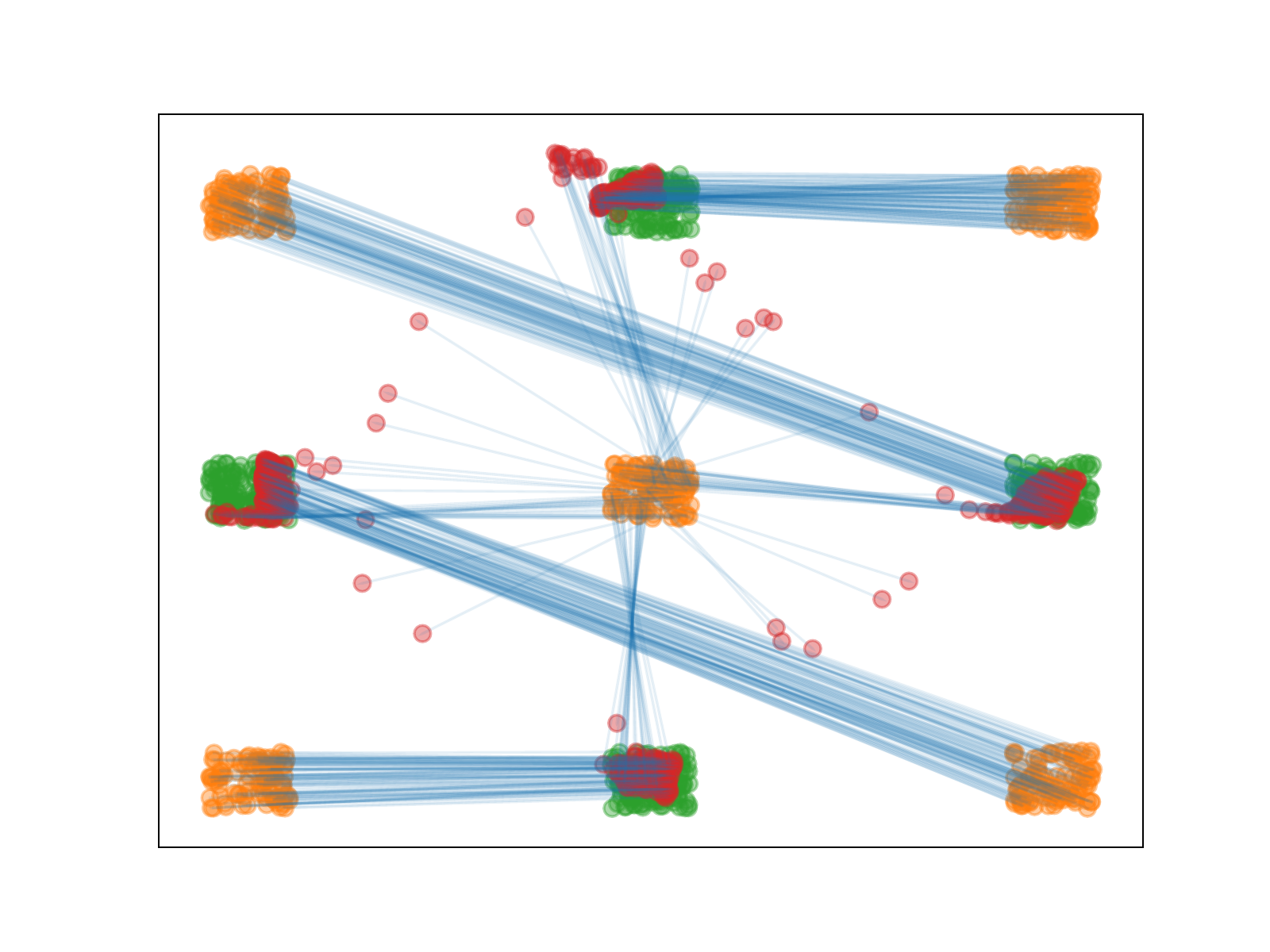}
			\label{fig:circle-map}
			& 	
\hspace{-0.35cm}
\includegraphics[trim=55 0 45 0, clip,width=0.24\hsize]{{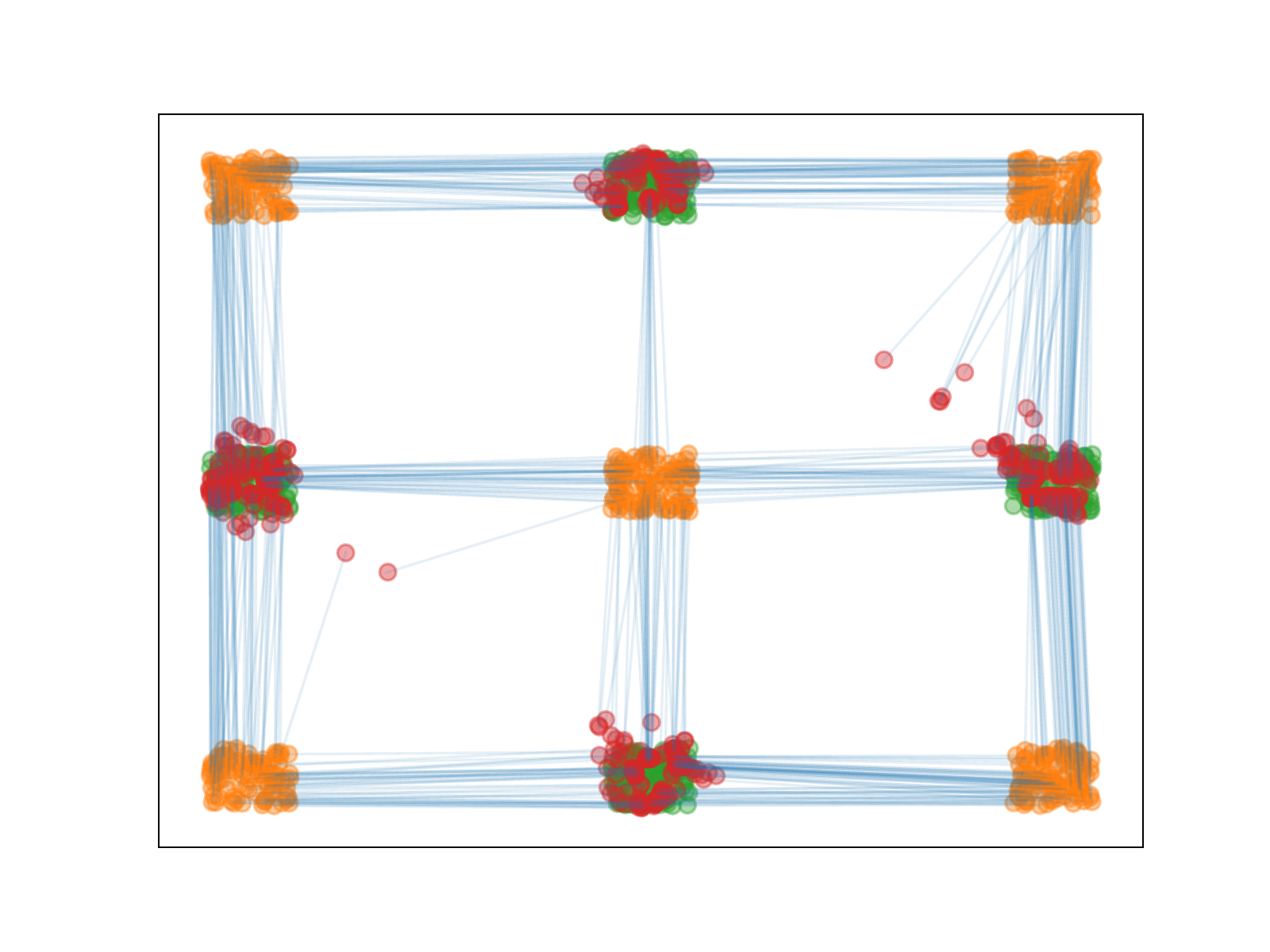}}
\label{fig:circle-map}
\vspace{-0.8cm}
\\ 
	\captionsetup[subfloat]{captionskip=-1.2pt}
		\hspace{-.65cm}
		\subfloat[ Barycentric-OT]{
			\includegraphics[trim=55 0 45 0, clip,width=0.24\hsize]{{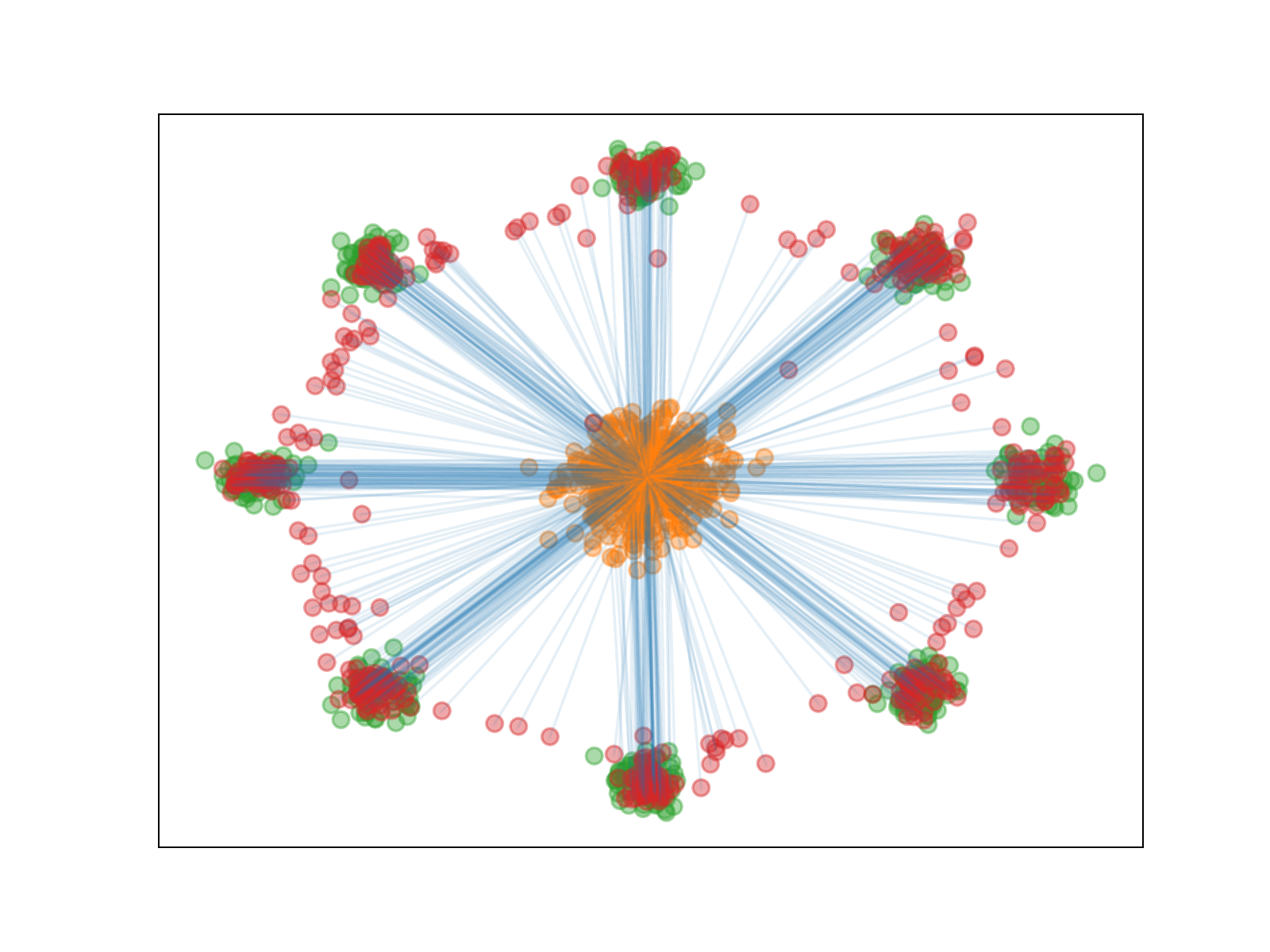}}
			\label{fig:bary_ot}
			} 
		& 	\hspace{-0.4cm}
			\captionsetup[subfloat]{captionskip=-1.2pt}
		\subfloat[W1-LP]{
			\includegraphics[trim=55 0 45 0, clip,width=0.24\hsize]{{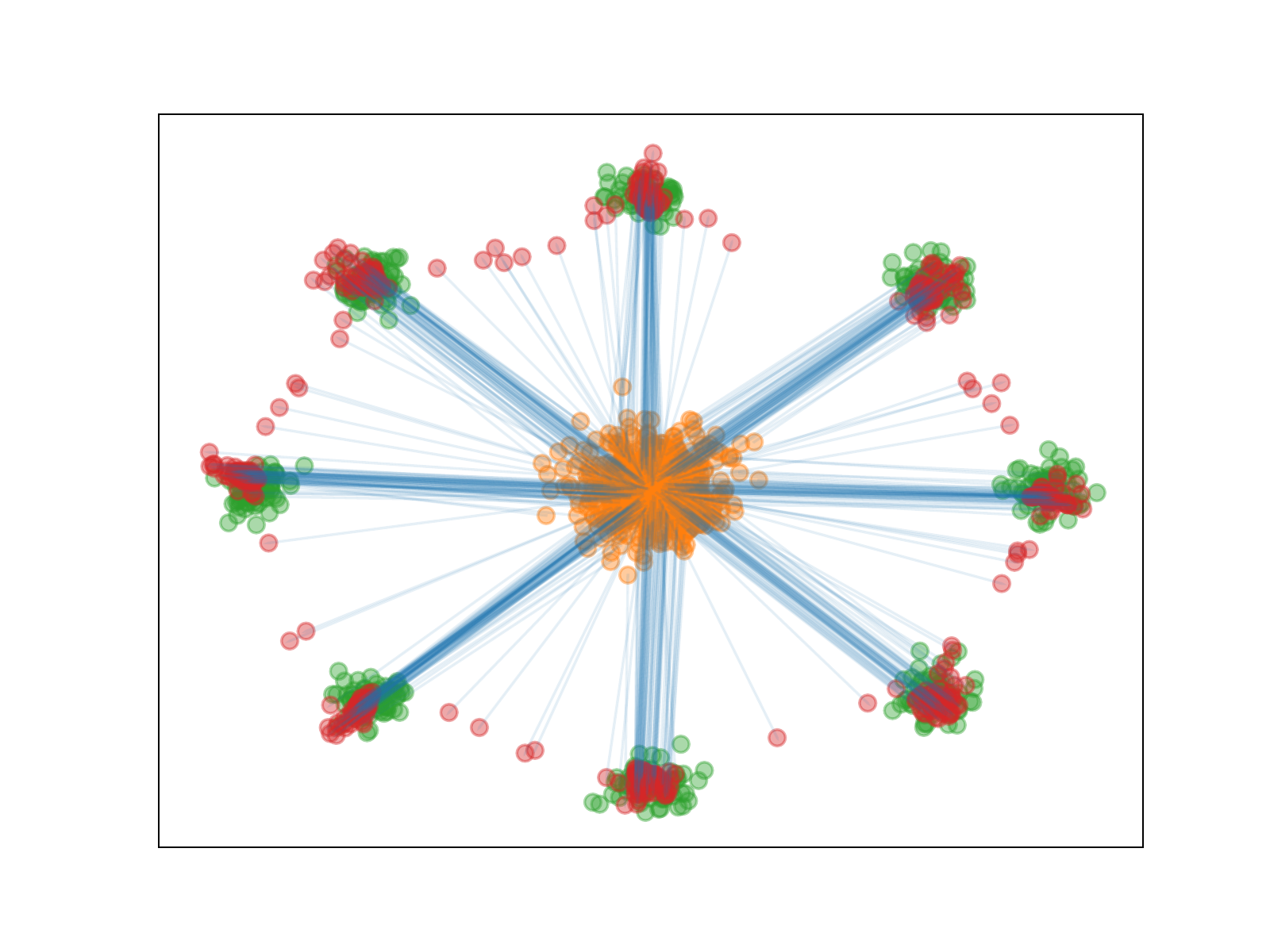}}\label{fig:w1-lp}
			} 
			&
		\hspace{-0.4cm}
			\captionsetup[subfloat]{captionskip=-1.2pt}
		\subfloat[W2GAN]{
			\includegraphics[trim=55 0 45 0, clip,width=0.24\hsize]{{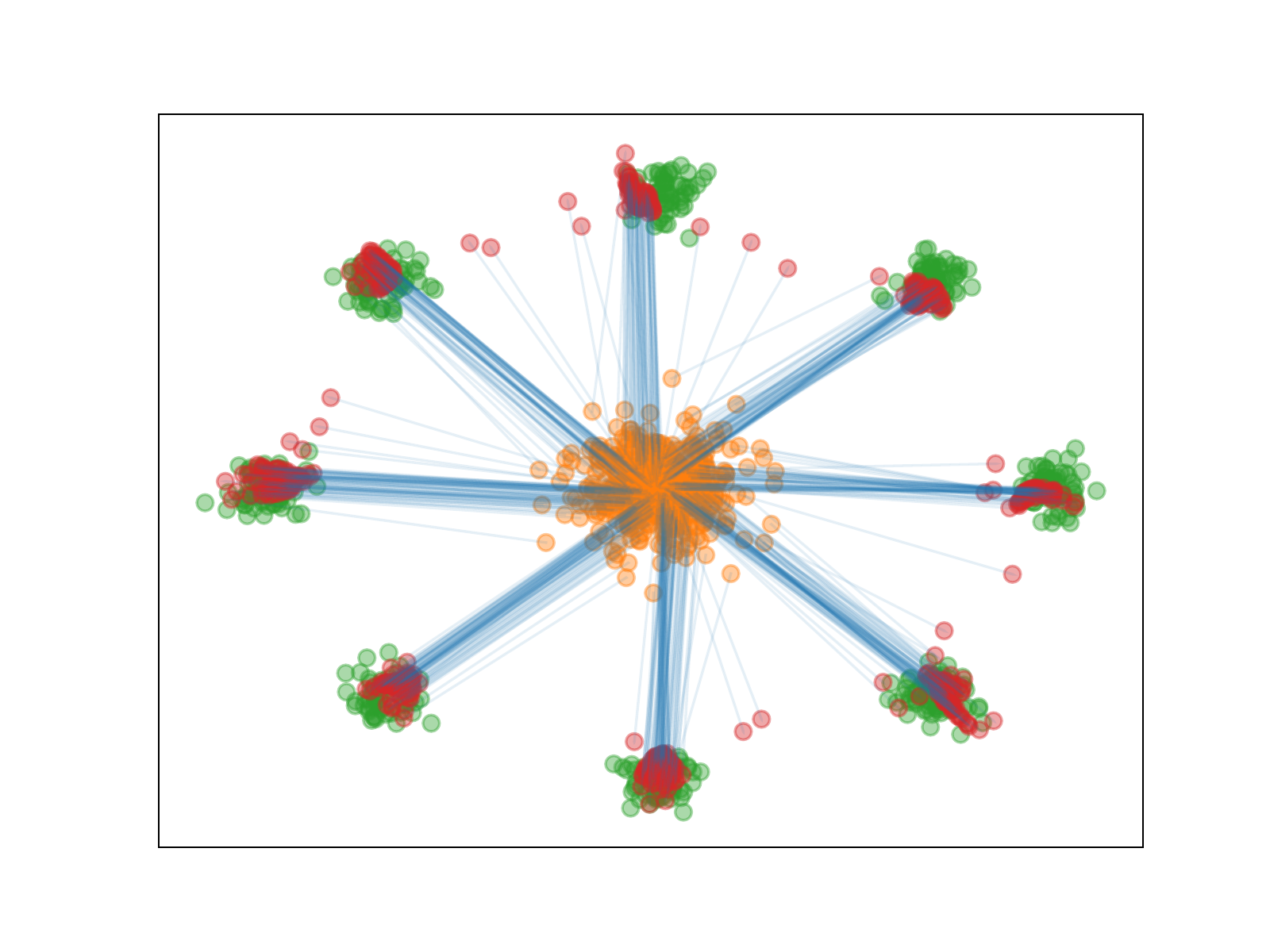}}
			\label{fig:w2gan}
			} 
			& 	
		\hspace{-0.4cm}
			\captionsetup[subfloat]{captionskip=-1.2pt}
		\subfloat[Our approach]{
			\includegraphics[trim=55 0 45 0, clip,width=0.24\hsize]{{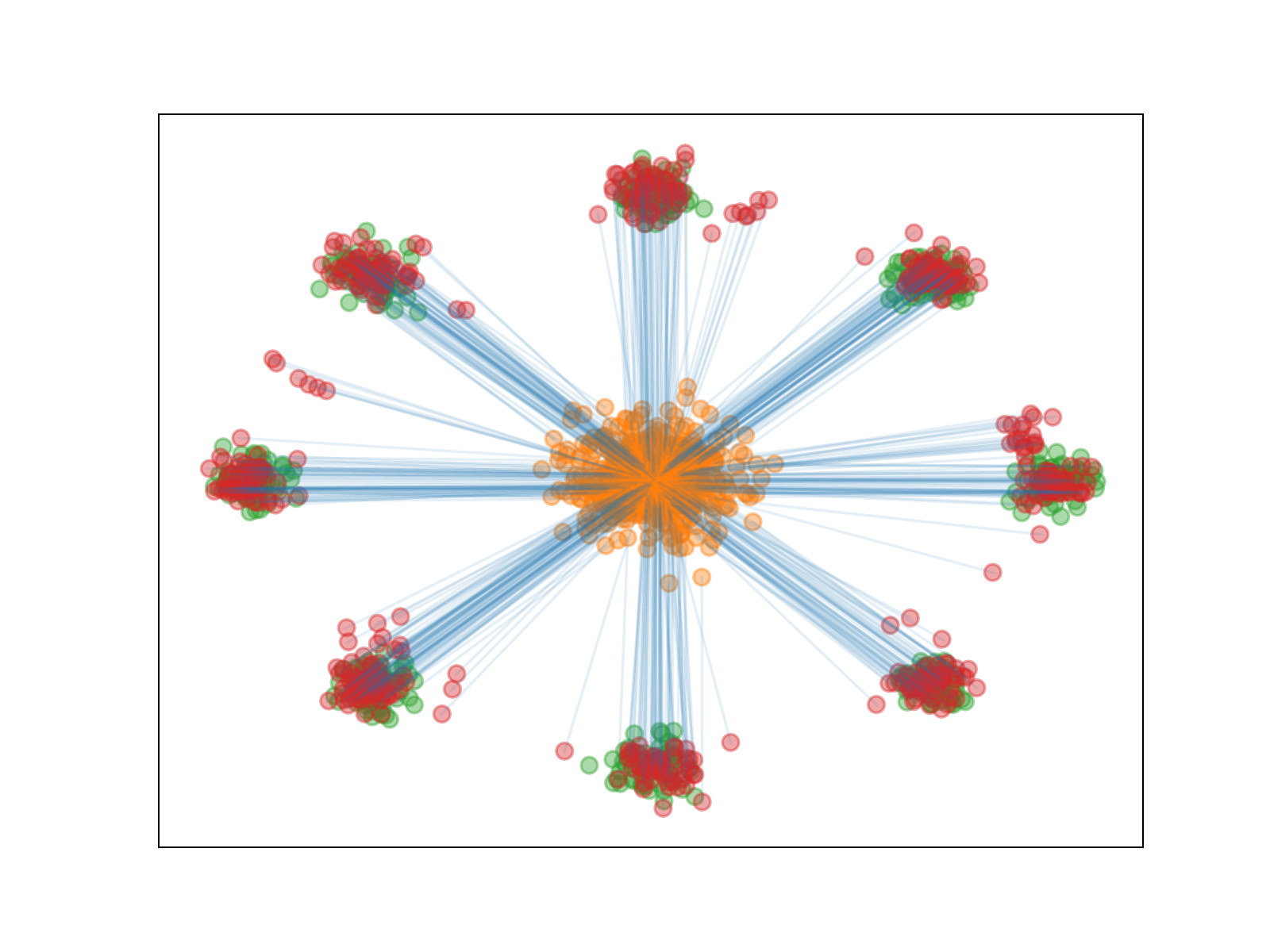}}
			\label{fig:ourapproach}}
	\end{tabular}
	\caption{The transport maps learned by various approaches on `Checker board' and `mixture of eight Gaussians' datasets. (a) Barycentric-OT \citep{seguy2017large}; (b) W1-LP \citep{petzka2017regularization}; (c) W2-GAN \citep{leygonie2019adversarial}; (d) Our approach (Algorithm~\prettyref{alg:W2}). The source distribution $Q$ is highlighted in orange, target distribution $P$ in green, the transported distribution $T_\#Q$ in red, and the transport map with  blue arrows.}
\label{fig:comparison}
\vspace{-0.4cm}
\end{figure*}

The proposed framework 
for learning the optimal transport 
provides a novel training method for deep generative models, where 
$(a)$ the generator is modeled as 
a gradient of a convex function and $(b)$ 
the minimax  optimization in \eqref{eq:approxW2}
(and more concretely, Algorithm \ref{alg:W2}) provides the training methodology. 
On the surface, Eq.~\eqref{eq:approxW2} resembles the minimax optimization of generative adversarial networks based on 
Wasserstein-1 distance \cite{arjovsky2017wasserstein}, called WGAN.
However, there are several critical differences making our approach attractive. 

First, because WGANs use optimal transportation distance only as a measure of distance, 
the learned generator map from the latent source to the target is arbitrary
and  sensitive to the initialization (see Figure~\ref{fig:W1-GAN}) \cite{jacob2018w2gan}. 
On the other hand, our proposed approach aims to find the {\em optimal} transport map and learns the same mapping regardless of 
the initialization (see Figure~\ref{fig:checker-board-OT0}).

Secondly, in a WGAN architecture~\cite{arjovsky2017wasserstein,petzka2017regularization}, 
the transport map (which is the generator) is represented with neural network that is a continuous mapping. 
Although, a discontinuous map can be approximated arbitrarily close with continuous 
neural networks, such a construction requires large weights making training unstable. 
On the other hand, through our proposed method, by representing the transport map with 
{\em gradient} of a neural network (equipped with ReLU type activation functions), 
we obtain a naturally  {\em discontinuous map}. 
As a consequence we have sharp transition from one part of the support to the other, whereas 
GANs (including WGANs) suffer from spurious probability masses that are not present in the target. 
This is illustrated in  Section~\ref{sec:exp-reg-OT}. 
The same holds for 
regularization-based methods for learning optimal transport~\cite{genevay2016stochastic,seguy2017large,leygonie2019adversarial}, 
where transport map is parametrized by continuous neural nets. 

\begin{remark}\label{rem:compare}
	\normalfont
	In a recent work, \citet{taghvaei20192} proposed to solve the semi-dual optimization problem~\eqref{eq:dual_convex_form} by representing the function $f$ with an ICNN and learning it using a stochastic optimization algorithm. However, each step of this algorithm requires computing the conjugate $f^*$ for all samples in the batch via solving a inner convex optimization problem for each sample which makes it slow and challenging to scale to large datasets. Further it is memory intensive as each inner optimization step requires a copy of all the samples in the dataset. In contrast, we represent the convex conjugate $f^\ast$ using ICNN and present a novel minimax formulation to learn it, in a scalable manner.

\end{remark}

\subsection{Stability analysis of the learned transport map}
\prettyref{thm:our_optim_result} establishes the consistency of our proposed optimization: if the objective \prettyref{eq:max-min} is solved exactly with a pair of functions $(f_0,g_0)$, then $\nabla g_0$ is the exact optimal transport map from $Q$ to $P$. In this section, we study the error in approximating the optimal transport map $\nabla g_0$, when the objective \prettyref{eq:max-min} is solved up to a small error. To this end, we build upon the recent results from \citet[Prop. 8]{hutter2019minimax} regarding the stability of optimal transport maps.
 
Recall that the optimization objective \prettyref{eq:max-min} involves a minimization and a maximization. For any pair $(f,g)$, let $\epsilon_1(f,g)$ denote the minimization gap and $\epsilon_2(g)$ denote the maximization gap, defined according to:
\begin{align}
\label{eq:geps}
\epsilon_1(f,g) &= \calV(f,g) - \inf_{\tilde{g} \in \cvx(Q)} \calV(f,\tilde{g}),\\
 \epsilon_2(f) &=  \sup_{\tilde{f}\in \cvx(P)} \inf_{\tilde{g} \in \cvx(Q)}\calV(\tilde{f},\tilde{g})  - \inf_{\tilde{g} \in \cvx(Q)} \calV(f,\tilde{g}) \nonumber
\end{align}
Then, the following theorem bounds the the error between $\nabla g$ and the optimal transport map $\nabla g_0$ as a function $\epsilon_1$ and $\epsilon_2$. We defer its proof to \prettyref{app:proof_stability}.

\begin{theorem}\label{thm:stability}
	Consider the optimization problem~\eqref{eq:max-min}. Assume $Q$ admits a density and let $\nabla g_0(\cdot)$ denote the optimal transport map from $Q$ to $P$. Then for any pair $(f,g)$ such that $f$ is $\alpha$-strongly convex, we have 
	\begin{align*}
	\|\nabla g - \nabla g_0 \|^2_{L^2(Q)} \leq\frac{2}{\alpha}(\epsilon_1(f,g)+\epsilon_2(f)),
	\end{align*}
	where $\epsilon_1$ and $\epsilon_2$ are defined in~\eqref{eq:geps}, and $\|\cdot\|_{L^2(Q)}$ denotes the $L^2$-norm with respect to measure $Q$. 	
\end{theorem}



\section{Experiments}
\label{sec:experiments}

In this section, first we qualitatively illustrate our proposed approach (see \prettyref{fig:comparison}) on the following two-dimensional synthetic datasets: (a) Checkerboard, (b) Mixture of eight Gaussians. We compare our method with the following three baselines: (i) Barycentric-OT \citep{seguy2017large}, (ii) W1-LP, which is the state-of-the-art Wasserstein GAN introduced by \cite{petzka2017regularization}, (iii) W2GAN \cite{leygonie2019adversarial}. Note that while the goal of W1-LP is not to learn the optimal transport map, the generator obtained at the end of its training can be viewed as a transport map. For all these baselines, we use the implementations (publicly available) of \citet{leygonie2019adversarial} which has the best set of parameters for each of these methods. In \prettyref{sec:exp-W1} and \prettyref{sec:exp-reg-OT}, we highlight the respective robustness and the discontinuity of our transport maps as opposed to other approaches. Finally, in \prettyref{sec:high-dim}, we show the effectiveness of our approach on the challenging task of learning the optimal transport map on a variety of synthetic and real world high-dimensional data. Full experimental details are provided in \prettyref{app:setup}.

\begin{figure*}[t]
\vspace{-0.3cm}
	\captionsetup[subfloat]{captionskip=-1pt}
	\centering
	\begin{tabular}{llll}
		\hspace{-.5cm}
		\subfloat[W1-LP: Trial $1$]{
			\includegraphics[trim=40 0 40 0, clip,width=0.25\hsize]{w1_checkerboard_2.pdf}
			\label{fig:checker-board-W1-1}} & 	
		\hspace{-0.5cm}
		\subfloat[W1-LP: Trial $2$]{
			\includegraphics[trim=40 0 40 0, clip,width=0.25\hsize]{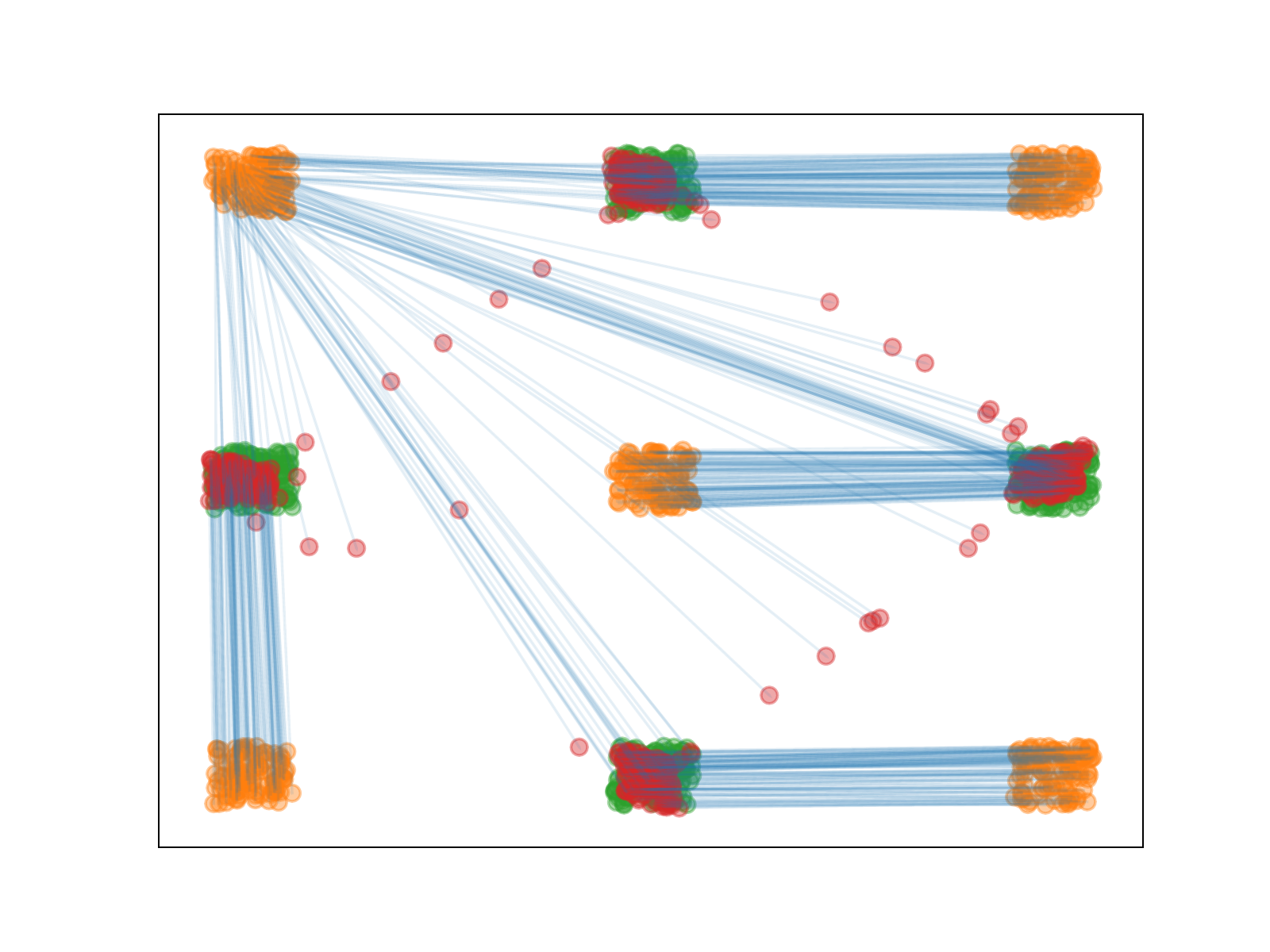}
			\label{fig:checker-board-W1-2}} &
		\hspace{-0.5cm}
		\subfloat[W2GAN: Trial $1$]{
			\includegraphics[trim=40 0 40 0, clip,width=0.25\hsize]{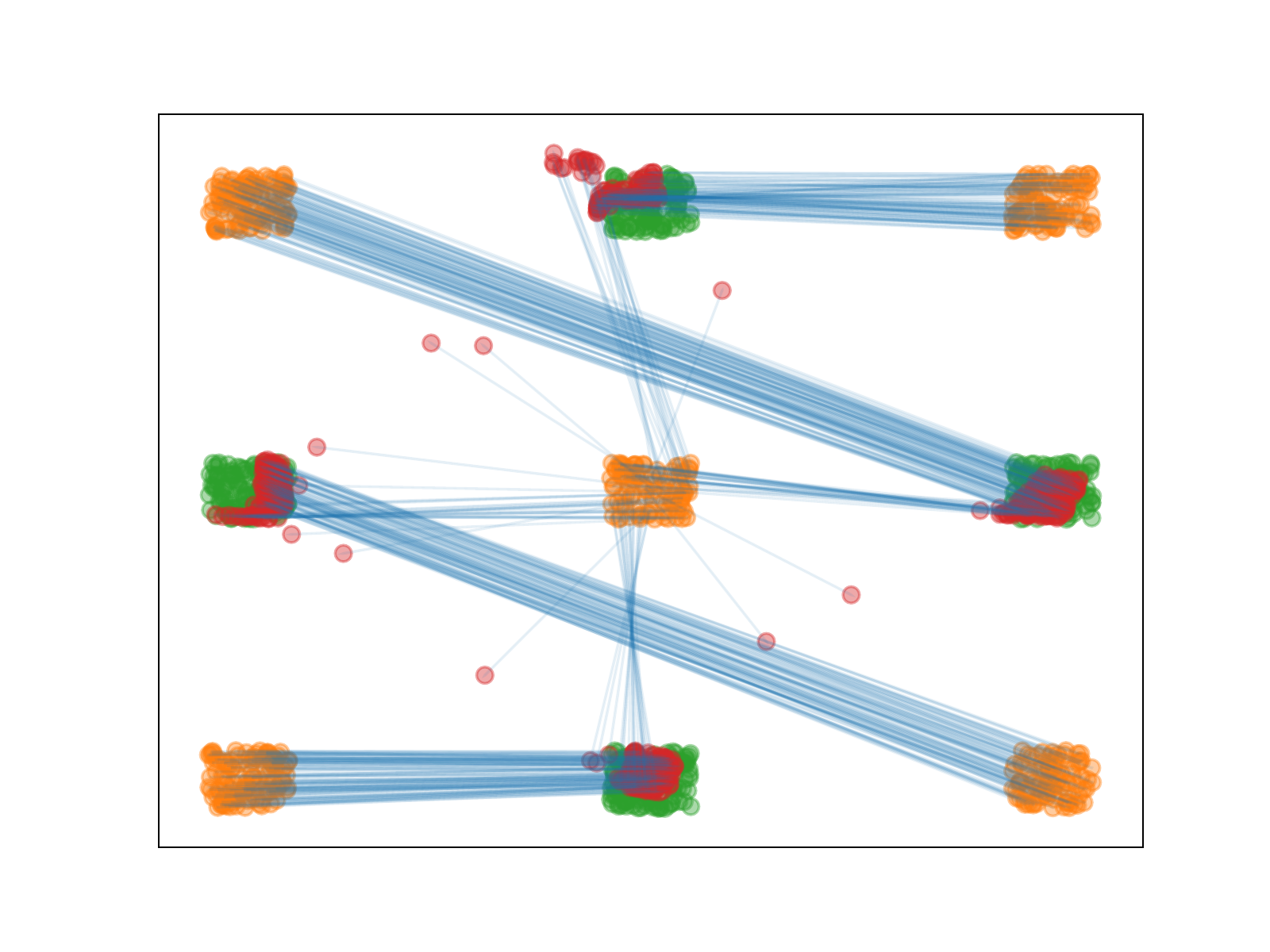} 
			\label{fig:checker-board-W2-1}} &
		\hspace{-0.5cm}
		\subfloat[W2GAN: Trial $2$]{
			\includegraphics[trim=40 0 40 0, clip,width=0.25\hsize]{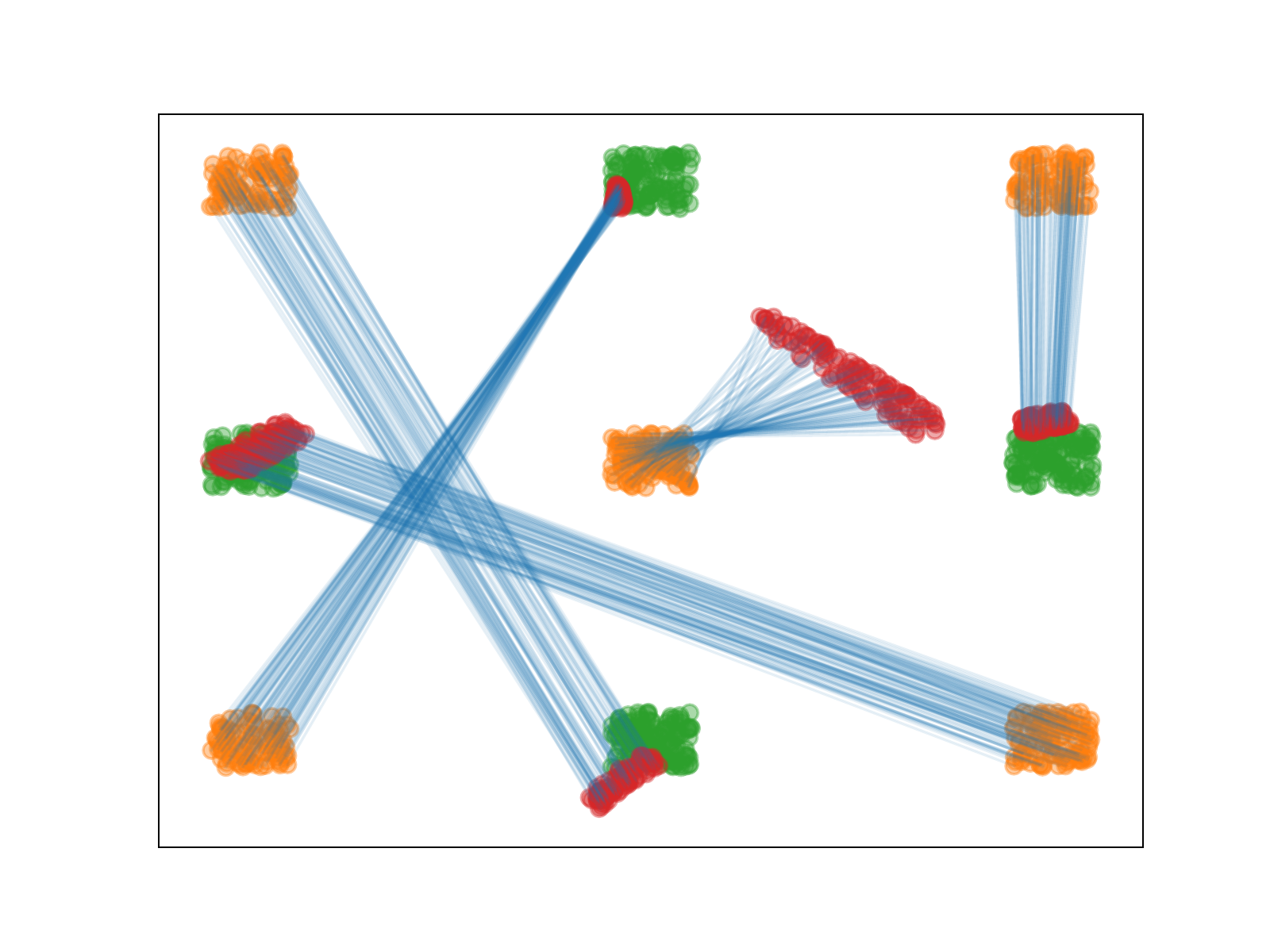} 
			\label{fig:checker-board-W2-2}} 
	\end{tabular}
	\caption{Transport maps learned by W1-LP \citep{petzka2017regularization} and W2GAN \citep{leygonie2019adversarial} under different random initialization.}
	\label{fig:W1-GAN}
	\vspace{-0.3cm}
\end{figure*}

{\bf Training methodology.}
We utilize our minimax formulation in \prettyref{eq:approxW2} to learn the optimal transport map. We parametrize the convex functions $f$ and $g$ using the same ICNN architecture (Figure~\ref{fig:ICNN}). Recall that to ensure convexity, we need to restrict all weights $W_\ell$'s to be non-negative (Assumption (i) in ICNN). We enforce it strictly for $f$, as the maximization over $g$ can be unbounded, making optimization unstable, whenever $f$ is non-convex. However,
we relax this constraint for $g$ (as permitted according to Remark~\ref{rem:relax}) and instead introduce a regularization term 
\begin{equation}
R(\theta_g) =\lambda  \sum_{ W_l \in \theta_g} \norm{\max (-W_l,0) }^2_F,
\end{equation}
where $\lambda>0$ is a regularization constant and the maximum is taken entry-wise for all the  weight parameters $\{W_l\} \subset \theta_g$. We empirically observe that this relaxation makes the optimization converge faster. 

For both the maximization and minimization updates 
in~\eqref{eq:approxW2}, we use  Adam \cite{kingma2014adam}. At each iteration, we draw a batch of samples from $P$ and $Q$ denoted by $\{X_i\}_{i=1}^{M}$ and $\{Y_j\}_{j=1}^{M}$ respectively. Then, we use the following objective for optimization which is an empirical counterpart of \eqref{eq:approxW2}:
\begin{equation}
\max_{\theta_f:  W_\ell  \geq 0,\forall\ell\in[L-1]} \min_{\theta_g}  \;\; J(\theta_f,\theta_g) +  R(\theta_g), 
\label{eq:J-samples}
\end{equation}
where $\theta_f,\theta_g$ are the parameters of  $f$ and $g$, respectively,  $W_\ell\geq0$ is an entry-wise constraint, and 
\begin{align*}
J(\theta_f,\theta_g) =  \frac{1}{M}\sum_{i=1}^{M} f(\nabla g(Y_i))   -  \langle Y_i, \nabla g(Y_i)\rangle  -f(X_i) .
\end{align*} 
This is summarized in Algorithm~\ref{alg:W2}. In the remainder of the paper, we interchangeably refer to Algorithm~\ref{alg:W2} as either `Our approach' or `Our algorithm'.

\begin{algorithm}
	\caption{The numerical procedure to solve the optimization problem~\eqref{eq:J-samples}.}
		\label{alg:W2}
			\begin{algorithmic}
	\STATE {\bfseries Input:} Source dist. $Q$, Target dist. $P$, Batch size $M$, Generator iterations $K$, Total iteratioins $T$ 
	\FOR{$t=1,\ldots, T$}
		\STATE Sample batch $\{X_i\}_{i=1}^{M} \sim P$ 
		\FOR{$k=1 ,\ldots,K$}
		 \STATE	Sample batch $\{Y_i\}_{i=1}^{M} \sim Q$
		 \STATE	Update $\theta_g$ to minimize~\eqref{eq:J-samples} using Adam method
		\ENDFOR
		 \STATE Update $\theta_f$ to maximize~\eqref{eq:J-samples} using Adam method
	 \STATE	Projection: $ w  \leftarrow \max(w,0)$, for all $w\in \{W^l\}\in \theta_f$ 
		\ENDFOR
	\end{algorithmic}
\end{algorithm}

\begin{remark}
\normalfont Note that the regularization term $R(\theta_g)$  is {\em data-independent} and does not introduce any bias to the optimization problem. For any convex function $f$, the minimizer of the problem~\eqref{eq:J-samples} is still a convex function $g$ as discussed in Remark~\ref{rem:relax}. We use this regularization to guide the algorithm towards neural networks that are convex.
\end{remark}
\subsection{Learning the optimal transport map}
\label{sec:exp_map}
As highlighted in \prettyref{fig:checker-board-OT0} and \prettyref{fig:ourapproach}, qualitatively, we observe that our proposed procedure indeed learns the optimal transport map on both the Checkerboard and Mixture of eight Gaussians datasets. In particular, our transport map is able to cut the continuous mass symmetrically and transport it to the nearest target support in both these examples. Also, Figure~\ref{fig:comparison} illustrates the qualitative difference of our approach compared to other approaches, in terms of non-optimality and existence of trailing dots. The existence of trailing dots is due to representing the transport map  with continuous neural networks, discussed in Section~\ref{sec:exp-reg-OT}.  

\subsection{Robustness of learning transport maps}
\label{sec:exp-W1}

In this section we numerically illustrate that the generator in W1-LP and W2GAN
finds arbitrary transport maps, and it is  sensitive to initialization as 
discussed in Section~\ref{sec:formulation}. 
This is in stark contrast with our proposed approach which finds the {\em optimal} transport independent of 
the initialization. 
We consider the previous Checkerboard example (Figure~\ref{fig:checker-board-data}) and train W1-LP and W2GAN with different random initializations.
The resulting transport maps for two different random trials are depicted in \prettyref{fig:checker-board-W1-1} and \prettyref{fig:checker-board-W1-2} for W1-LP, and \prettyref{fig:checker-board-W2-1} and \prettyref{fig:checker-board-W2-2} for W2-GAN. 
In addition to the fact that the learned transport map is very sensitive to initializations,  the quality of the samples generated by thus trained models are also sensitive. This is a major challenge in training GANs \cite{lin2018pacgan}.


\subsection{Learning discontinuous transport maps}
\label{sec:exp-reg-OT}

The power to represent a discontinuous transport mapping is 
what fundamentally sets our proposed method apart from  
the existing approaches, as discussed in Section~\ref{sec:formulation}. 
Two prominent approaches for learning transport maps are 
generative adversarial networks \cite{arjovsky2017wasserstein,petzka2017regularization} 
and regularized optimal transport \cite{genevay2016stochastic,seguy2017large}. 
In both cases, the transport map is modeled by a standard neural network with finite depth and width, which is a continuous function. 
As a consequence, 
continuous transport maps suffer from unintended and undesired spurious probability mass that 
connects disjoint supports of the target probability distribution. 

First, standard GANs including 
the original GAN \cite{goodfellow2014generative} and  
variants of WGAN \cite{arjovsky2017wasserstein,gulrajani2017improved,wei2018improving}
all suffer from spurious probability masses. 
Even those designed to tackle such spurious probability masses, like PacGAN \cite{lin2018pacgan},  
cannot overcome the barrier of continuous neural networks. 
This suggests that fundamental change in the architecture, like the one we propose, is necessary. \prettyref{fig:w1-lp} illustrates the same scenario for the transport map learned through the WGAN framework.  
We can observe the trailing dots of spurious probability masses, 
resulting from undesired continuity of the learned transport maps. 

Similarly, regularization methods to approximate optimal transport maps, explained in Section~\ref{sec:background}, 
suffer from the same phenomenon. 
Representing a transport map with an inherently continuous function class results in 
spurious probability masses connecting disjoint supports. \prettyref{fig:bary_ot}, corresponding to Barycentric-OT, illustrates those trailing dots of spurious masses 
for the  learned transport map from algorithm introduced in \citet{seguy2017large}. We also observe a similar phenomenon with \citet{leygonie2019adversarial} as illustrated in \prettyref{fig:w2gan}.

On the other hand, 
we represent the transport map with the {\em gradient} of 
a neural network (equipped with non-smooth ReLU type activation functions). 
The resulting transport map can naturally represent  discontinuous transport maps, 
as illustrated in 
Figure~\ref{fig:checker-board-OT} and  \prettyref{fig:ourapproach}.  
The vector field of the learned transport map in Figure~\ref{fig:g-vector-field} clearly shows 
the discontinuity of the learned optimal transport. 



\subsection{High dimensional experiments}
\label{sec:high-dim}

\begin{figure*}[t]
	\captionsetup[subfloat]{captionskip=7pt}
	\centering
	\begin{tabular}{cccc}
		\hspace{-.8cm}
		\subfloat[Estimated distance]{
					\includegraphics[trim=0 -4 0 0, clip,width=0.24\hsize]{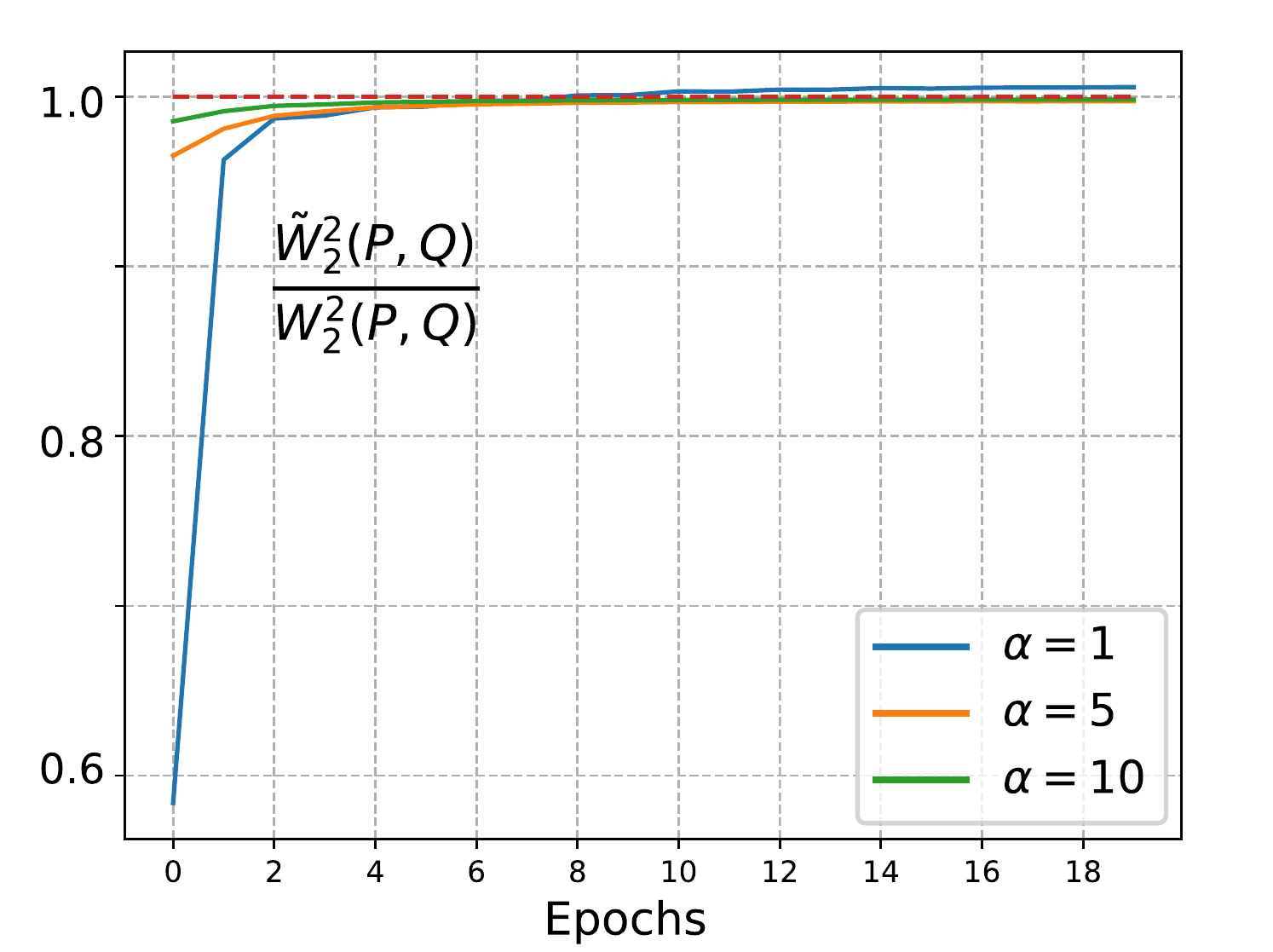}
			\label{fig:w2-convergence}}
						 & 	
		\hspace{-0.6cm}
		\subfloat[High-dim Gaussian to $2$-dim mixture]{
					\includegraphics[trim=38 0 20 0, clip, width=0.355\hsize]{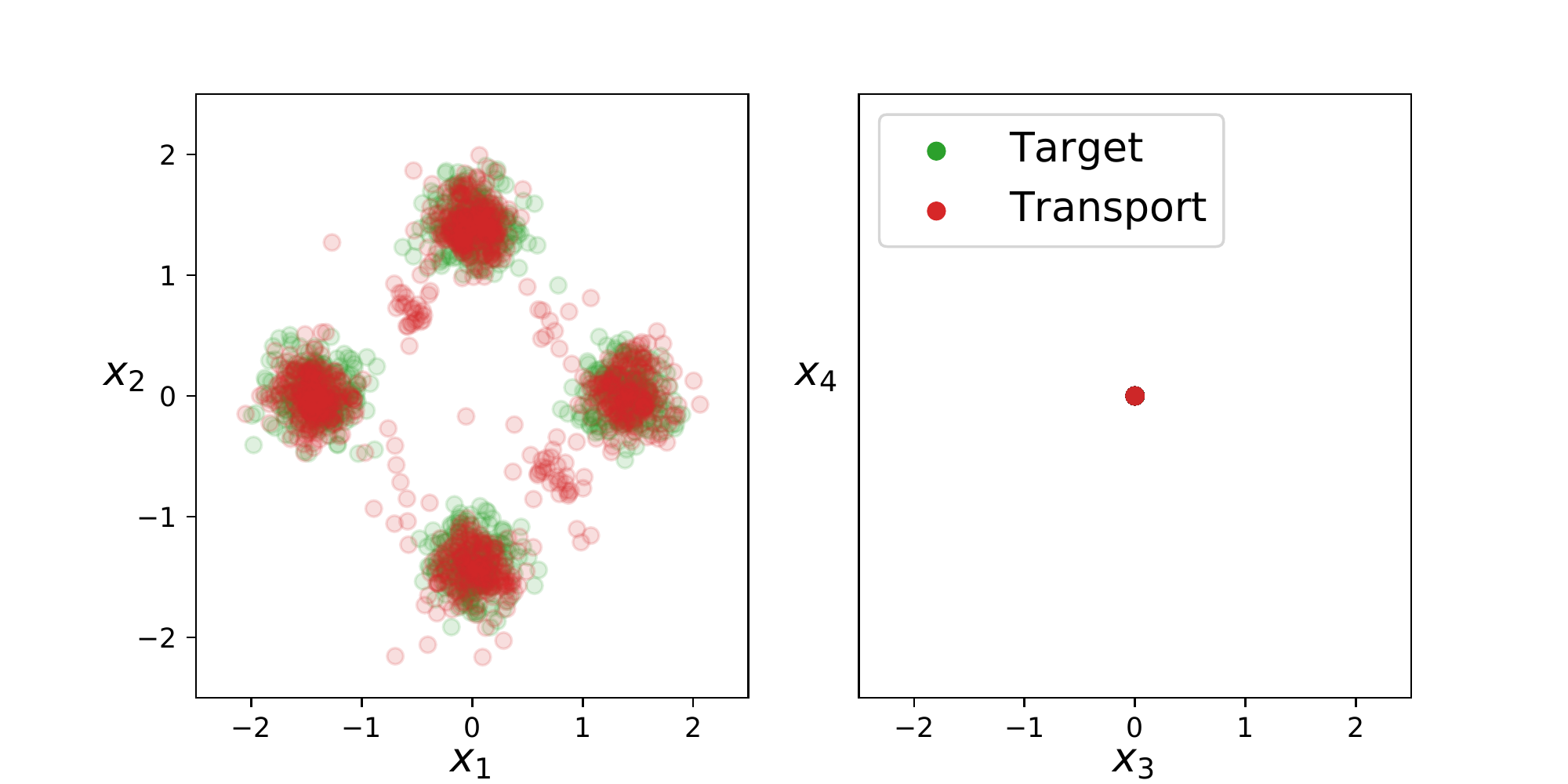}
			\label{fig:mixture-high-dim}}
						 &
		\hspace{-.5cm}
		\subfloat[Source: First $5$ digits]{
			\includegraphics[width=0.18\hsize, height=3.0cm]{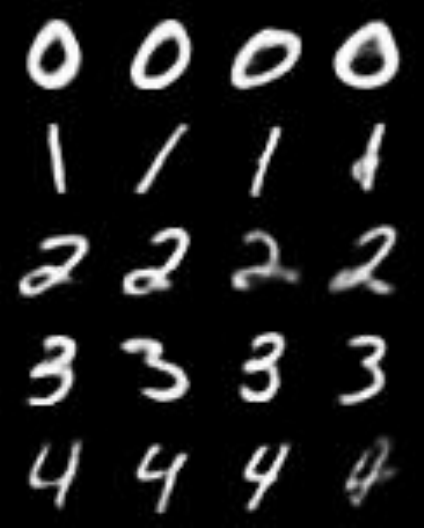} 
			\label{fig:mnist-firstfive}} 
			&
		\hspace{-0.5cm}
		\subfloat[Transported samples]{
			\includegraphics[width=0.18\hsize, height=3.0cm, clip]{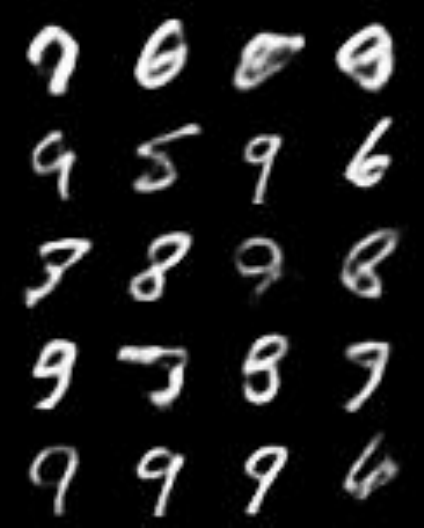}   
			\label{fig:mnist-lastfive}} 
	\end{tabular}
	\caption{ Numerical results on high-dimensional experiments of Section~\ref{sec:high-dim}: (a) Convergence of our estimated $W_2$ distance to the actual value when transporting $\calN(0,I_d)$ to $\calN(\alpha \mathbbm{1},I_d)$ where $d=784$; (b) Transporting a $784$-dim Gaussian to a $2$-dim Gaussian mixture embedded in $784$-dim space; (c) Samples from the source distribution corresponding to first five MNIST digits, embedded into $16$-dim. feature space. (d) Image of the samples under the learned optimal transport map, where the target distribution is the last five digits.}
 \label{fig:all_high_dim}
\end{figure*}

We consider the challenging task of learning optimal transport maps on high dimensional distributions. In particular, we consider both synthetic and real world high dimensional datasets and provide quantitative and qualitative illustration of the performance of our proposed approach. 

{\bf Gaussian to Gaussian.} Source distribution $Q=\calN(0,I_d)$ and target distribution $P=\calN(\mu,I_d)$, for some fixed $\mu\in\reals^d$ and $d=784$. The mean vector $\mu=\alpha (1,\ldots,1)^\top$ for some parameter $\alpha>0$. Because both distributions are Gaussian, the optimal transport map is explicitly known: $T^\ast(x)=x+\mu$ and hence $W_2^2(P,Q)=\norm{\mu}^2/2=\alpha^2 d/2$. In \prettyref{fig:w2-convergence}, we compare our estimated distance $\tilde{W}_2^2(P,Q)$, defined in~\eqref{eq:approxW2}, with the exact value $W_2^2(P,Q)$, as the training progresses for various values of $\alpha \in \{1,5,10\}$. Intuitively, learning is more challenging when $\alpha$ is larger. Further, error in learning the optimal transport map, quantified with the metric $\norm{\mu_{T(Q)}-\mu}^2$, where $\mu_{T(Q)}$ is the mean of the transported distribution $T_\#Q$, is reported in Table~\ref{tab:gauss_gauss_table}.


\begin{table}[h]
\caption{The error between the mean of transported and that of the target distributions. The source and target are $728$-dim. Gaussians.}
\tiny
\label{tab:gauss_gauss_table}
\begin{center}
\begin{sc}
\begin{tabular}{cccc}
Metric & $\alpha=1$ & $\alpha=5$ & $\alpha=10$ \\
\midrule
$\norm{\mu_{T(Q)}-\mu}^2$    & $ 0.19\pm 0.015$ & $13.95 \pm 1.45$ & $29.05 \pm 5.16$ \\
\midrule
\hspace{-1.7em} $100 \cdot (\norm{\mu_{T(Q)}-\mu}/\norm{\mu})^2$    & $ 0.02\pm 0.001$ & $0.07 \pm 0.005$ & $0.04 \pm 0.006$ \\
\bottomrule
\end{tabular}
\end{sc}
\end{center}

\end{table}


{\bf High-dim. Gaussian to low-dim. mixture.} Source distribution $Q$ is standard Gaussian $\calN(0,I_d)$ with $d=784$, and the target distribution $P$ is a mixture of four Gaussians that lie in in the two-dimensional subspace of the high-dimensional space $\real^d$,  \ie the first two components of the random vector $X\sim P$ is mixture of four Gaussians, and the rest of the components are zero.  The projection of the learned  optimal transport map onto the first four components is depicted in Figure~\ref{fig:mixture-high-dim}. As illustrated in the left panel of \ref{fig:mixture-high-dim}, our transport map correctly maps the source distribution to the mixture of four Gaussians in the first two components.  And  it maps the rest of the components to zero, as highlighted by a red blob at zero in the right panel.

{\bf MNIST $\{0,1,2,3,4\}$ to MNIST $\{5,6,7,8,9\}$.} We consider the standard MNIST dataset~\cite{mnist} with the goal of learning the optimal transport map from the set of images corresponding to first five digits~$\{0,1,2,3,4\}$ to the last five digits~$\{5,6,7,8,9\}$. To achieve this, we embed the images into the a space where the Euclidean norm $\|\cdot\|$ between the embedded images is meaningful. This is in alignment with the reported results in the literature for learning the $L_2$-optimal transport map~\citep[Sec. 4.1]{yang2019potential}. We consider the embeddings into a $16$-dimensional latent feature space given by a pre-trained Variational Autoencoder (VAE). We simulate our algorithm on this feature space. 
The results of the learned transport map are depicted in \prettyref{fig:all_high_dim}. Figure~\ref{fig:mnist-firstfive} presents samples from the source distribution and Figure~\ref{fig:mnist-lastfive} illustrates the source samples after transportation under the learned optimal transport map. We observe that the digits that look alike are coupled via the optimal transport map, e.g. $1\to9$, $2\to8$, and $4\to9$. 

\begin{figure}[h]
	\centering 
	\includegraphics[width=0.25\textwidth]{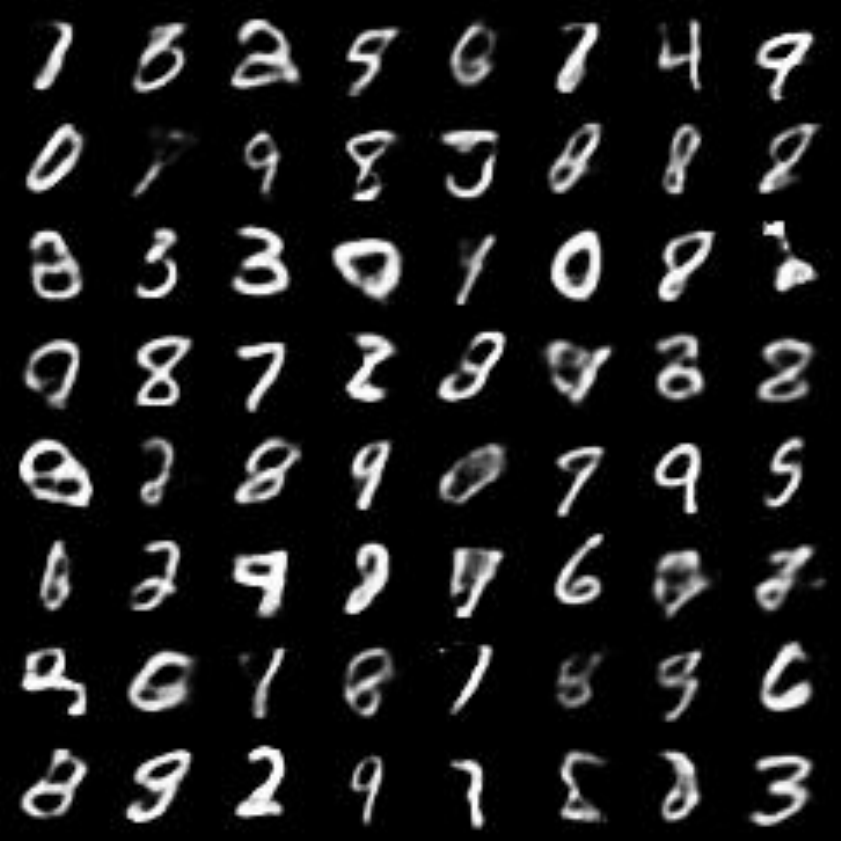}
	\caption{MNIST like samples generated by the learned optimal transport map from Gaussian source distribution in feature space.}
		\label{fig:gauss-mnist}
\end{figure}

{\bf Gaussian to MNIST.} The source is $16$-dimensional standard Gaussian distribution, and the target is the $16$-dimensional latent embeddings of all the MNIST digits. The MNIST like samples  that are generated from the learned optimal transport map are depicted in~\prettyref{fig:gauss-mnist}. 

These experiments serve as a proof of concept that the algorithm scales to high-dimensional setting and real-world dataset. We believe that further improvements on the performance of the proposed algorithm requires careful tuning of hyper-parameters which
takes time to develop (similar to initial WGAN) and is a subject of ongoing work.

\vspace{-10pt}
\section{Conclusion}
We presented a novel minimax framework to learn the optimal transport map under $W_2$-metric. Our framework is in contrast to regularization-based approaches, where the constraint of the dual Kantorovich problem is replaced with a penalty term. Instead, we represent the dual functions with ICNN, so that the constraint is automatically satisfied. Further, the transport map is expressed as gradient of a convex function, which is able to represent discontinuous  maps. We believe that our  framework paves way for bridging the optimal transport theory and practice.

\bibliographystyle{icml2020}
\bibliography{ref}

\clearpage
\onecolumn
\appendix

\section{Proof of Theorem~\ref{thm:our_optim_result}}
\label{app:proof_theorem}

Define $V_f(g) \triangleq \Expect_Q[\langle Y,\nabla g(Y)  \rangle - f(\nabla g(Y))]$. The main step of the proof is to show that $\sup_{g \in \cvx(Q)}  V_f(g) = \Expect_Q[f^*(Y)]$. Then the conclusion follows from \eqref{eq:dual_convex_form}. To prove this, note that for all $g \in \cvx(Q)$, we have
\[
\langle y, \nabla g(y) \rangle - f(\nabla g(y))  \leq \langle y, \nabla f^*(y) \rangle - f(\nabla f^*(y))   = f^*(y), 
\]
for all $y \in \real^d$ such that $g$ and $f^*$ are differentiable at $y$. We now claim that both $g$ and $f^*$ are differentiable $Q$-almost everywhere (a.e). If the claim is true, upon taking the expectation w.r.t $Q$: 
\[V_f(g) \leq V_f(f^*)  =  \Expect_Q[f^*(Y)],\quad \forall g \in \cvx(Q)\]
and the inequality is achieved with $g=f^*$. Now we prove the claim as follows: Since $\int g \ud Q < \infty$, we have $Q(g=\infty)=0$. Thus $Q(\text{Dom}(g))=1$, where $\text{Dom}(g)$ is the domain of the function $g$.  Moreover, $Q(\text{Int}(\text{Dom}(g)) =1$, where $\text{Int}(\cdot)$ denotes the interior, because the boundary has $Q$-measure zero ($Q$ has a density). Since $g$ is convex, it is differentiable on $\text{Int}(\text{Dom}(g))$ except at points of Lebesgue measure zero which have $Q$-measure zero too. Therefore, $g$ is $ Q$-a.e differentiable.  Similar arguments hold for $f^*$.

\section{Proof of Theorem~\ref{thm:stability}}
\label{app:proof_stability}
The  proof follows from the bounds 
\begin{subequations}
\begin{align}
\|\nabla g - \nabla f^* \|^2_{L^2(Q)} \leq  \frac{2}{\alpha} \epsilon_1\label{eq:bound1},\\
\|\nabla f^* - \nabla g_0 \|^2_{L^2(Q)} \leq  \frac{2}{\alpha} \epsilon_2\label{eq:bound2},
\end{align}
\end{subequations}
and using the triangle inequality. The proof for the first bound is as follows. 
If $f$ is $\alpha$-strongly convex, then $f^*$ is $\frac{1}{\alpha}$ smooth. By definition of smoothness, 
\begin{equation*}
f^*(z) \leq f^*(y) + \langle \nabla f^*(y), z-y \rangle  + \frac{1}{2\alpha}\|z-y\|^2 \triangleq h_y(z),\quad \forall y,z \in \real^d,
\end{equation*}
where $h_y(z)$ is defined to be the quadratic function of $z$ that appears on the right-hand side of the inequality. From  $f^*(z)\leq h_y(z)$, it follows that the convex conjugate  $f(x) \geq h^*_y(x)$. As a result,  
\begin{equation}
f(x) \geq h^*_y(x)= -f^*(y) + \langle y,x \rangle + \frac{\alpha}{2}\|x-\nabla f^*(y)\|^2,\quad \forall x,y\in \real^d.
 \label{eq:inequality}
\end{equation}
We use this inequality to control the optimality gap $\epsilon_1(f,g)$:
\begin{align*}
\epsilon_1(f,g)&=\calV(f,g) - \inf_{\tilde{g}} \calV (f,\tilde{g})  \\&= \calV(f,g)-\calV(f,f^*)  \\&= \Expect_Q   [f^*(Y) - \langle Y,\nabla g(Y) \rangle  + f(\nabla g(Y))  ]
 \\&\geq  \frac{\alpha}{2} \Expect_Q[\| \nabla g(Y) - \nabla f^*(Y)\|^2 ],
\end{align*}
where the last step follows from~\eqref{eq:inequality}, with $x=\nabla g(y)$. This concludes the proof of the bound~\eqref{eq:bound1}. It remains to prove~\eqref{eq:bound2}.  To this end, note that the optimality gap $\epsilon_2(f)$ is given by
\begin{align*}
\epsilon_2(f)&=\calV(f_0,g_0) -  \inf_{\tilde g} \calV(f,\tilde{g}) \\&= \calV(f_0,f^*_0) - \calV(f,f^*) \\&= - (\Expect_P[f_0(X)]+ \Expect_Q   [f^*_0(Y)]) + (\Expect_P[f(X)]+ \Expect_Q   [f^*(Y)])  
\\&=  - \Expect_Q [f_0(\nabla f^*_0(Y)) + f^*_0(Y)] + \Expect_Q [f(\nabla f^*_0(Y)) + f^*(Y)] 
\\&=    - \Expect_Q [\langle Y, \nabla f^*_0(Y)\rangle ] +\Expect_Q [f(\nabla f^*_0(Y)) + f^*(Y)]
\end{align*}
Using the inequality~\eqref{eq:inequality} with $x=\nabla f^*_0(y)$ yields:
\begin{align*}
\epsilon_2(f)
&\geq  \frac{\alpha}{2} \Expect_Q[| \nabla f^*_0(Y) - \nabla f^*(Y)|^2 ] 
\end{align*}
concluding~\eqref{eq:bound2} noting that $f^*_0=g_0$. 
\section{Experimental set-up}
\label{app:setup}

\subsection{Two-dimensional experiments}

{\bf Datasets.} We use the following synthetic datasets: (i) Checkerboard, and (ii) Mixture of eight Gaussians. For the Checkerboard dataset, the source distribution $Q$ is the law of the random variable $Y=X + Z$, where $X \sim \mathrm{Unif}(\{(0,0), (1,1),(1,-1),(-1,1),(-1,-1)\})$ and $Z \sim \mathrm{Unif}([-0.5,0.5] \times [-0.5, 0.5])$. Similarly, $P$ is the distribution of random variable $Y=X + Z$, where $X \sim \mathrm{Unif}(\{(0,1), (0,-1),(1,0),(-1,0)\})$ and $Z \sim \mathrm{Unif}([-0.5,0.5] \times [-0.5, 0.5])$. Note that $\mathrm{Unif}(B)$ denotes the uniform distribution over any set $B$. For the mixture of eight Gaussians dataset, we have $Q=\calN(0,I_2)$ and $P$ is the law of random variable $Y$, where $Y=X+Z$ with $X \sim \mathrm{Unif}(\{(1,0), (\frac{1} {\sqrt{2}},\frac{1} {\sqrt{2}})\},(0,1),(\frac{-1} {\sqrt{2}},\frac{1} {\sqrt{2}}), (-1,0),(\frac{-1} {\sqrt{2}},\frac{-1} {\sqrt{2}}), (0,-1), (\frac{1} {\sqrt{2}},\frac{-1} {\sqrt{2}}) \}) $ and $Z \sim \calN(0,0.5I_2)$.

{\bf Architecture details.}
For our Algorithm~\ref{alg:W2}, we parametrize both the convex functions $f$ and $g$ by ICNNs. Both these ICNN networks have equal number of nodes for all the hidden layers 
followed by a final output layer. 
We choose 
a square of leaky ReLU function, i.e $\sigma_0(x) = \left(\text{max}(\beta x,x) \right)^2$ with a small positive constant $\beta$ as 
the {\em convex} 
activation function for the first layer $\sigma_0$. 
For the remaining layers, we use the leaky ReLU function, i.e $\sigma_l(x) = \text{max}(\beta x,x)$ for $l=1,\ldots,L-1$, 
as the {\em monotonically non-decreasing and convex} activation function.  
Note that the assumptions (ii)-(iii) of the ICNN are satisfied.
In all of our experiments, we set the parameter $\beta = 0.2$. In some of the experiments as explained below, we chose the SELU activation function which also obeys the convexity assumptions.

For the three baselines, Barycentric-OT, W1-LP, and W2GAN, we use the implementations of \citet{leygonie2019adversarial}, made publicly available at \url{https://github.com/jshe/wasserstein-2}. For all these methods, we use the default settings of hyperparameters which were fixed to be the best values from the respective papers. Further, for a fair comparison we allow the number of parameters in each of these baselines to be larger than ours; in fact, for W2GAN and Barycentric-OT, the default number of neural network parameters is much larger than ours.

{\bf Hyperparameters.} For reproducibility, we provide the details of the numerical experiments for each of the figures. For the Checkerboard dataset in \prettyref{fig:comparison} (same as \prettyref{fig:checker-board-OT0}),  
we run Algorithm~\ref{alg:W2} with the following parameters: For both the ICNNs $f$ and $g$, we set the hidden size $m=64$, number of layers $L=4$, regularization constant $\lambda=1.0$, Leaky ReLU activation and for training we use batch size $M=1024$, learning rate $10^{-4}$, generator iterations $K=10$, total number of iterations $T=10^5$, and the Adam optimizer with $\beta_1=0.5$, and $\beta_2=0.9$. For each of the baselines, the following are the values of the parameters: (a) Barycentric-OT: $3$ ($1$ corresponding to the dual stage and the rest for the map step) neural networks each with $m=128, L=3, M=512, T=2\times 10^5$ and $l_2$-entropy penalty, (b) W1-LP: Both the discriminator and the generator neural networks with $m=128, L=3, K=5$ and  $M=512, T=2\times 10^5$, and (c) W2GAN: $3$ neural networks ($1$ corresponding to the generator whereas the remaining are for two functions in the dual formulation \prettyref{eq:dual_form}) each with $m=128, L=3, K=5, M=512, T=2\times 10^5$. W2GAN also uses six additional regularization terms which set to default values as provided in the code. Also, all these baselines use ReLU activation and Adam optimizer with $\beta_1=0.9$ and $\beta_2=0.990$ and the learning rate for generator parameters being $0.0001$ and $0.0005$ for the rest. For the mixture of eight Gaussians dataset, we use the same parameters except batch-size $M=256$, whereas all the baselines use the same parameters as the above setting. Also, for the multiple trials in \prettyref{fig:W1-GAN} for W1-LP and W2GAN, we use the above parameters but with a different random initialization of the neural network weights and biases.

\subsection{High dimensional experiments}
\label{app:high-dim-exp}

{\bf Gaussian to Gaussian.} Source distribution $Q=\calN(0,I_d)$ and target distribution $P=\calN(\mu,I_d)$, for some fixed $\mu\in\reals^d$ and $d=784$. The mean vector $\mu=\alpha (1,\ldots,1)^\top$ with $\alpha \in \{1,5,10\}$. For both the ICNNs $f$ and $g$, we have $d=784, m=1024, L=3$, Leaky ReLU activation, batch size $M= 60$, $K=16$, $\lambda=0.1$, $T=40,000$, Adam optimizer with $\beta_1=0.5$ and $\beta_2 =0.99$, learning rate decay by a factor of $0.5$ for every $2,000$ iterations. Note that in \prettyref{fig:w2-convergence}, $1$ epoch corresponds to $1000$ iterations.

{\bf High-dim. Gaussian to low-dim. mixture.} Source distribution $Q=\calN(0,I_d)$ with $d=784$. The target distribution is a mixture of four Gaussians $P=\sum_{i=1}^4\frac{1}{4}\calN(\mu_i,\Sigma )$, where  $\mu_i=(\pm 1.4,\pm 1.4, 0 , \ldots, 0) \in \real^{784}$ and $\Sigma=\text{diag}(0.2,0.2,0,\ldots,0)$. For both the ICNNs $f$ and $g$, we have $d=784, m=1024, L=3$, Leaky ReLU activation, batch size  $M=60$, $K=25$, $\lambda=0.01$, Adam optimizer with $\beta_1=0.5$ and $\beta_2 =0.99$, learning rate decay by a factor of $0.5$ for every two epochs. The algorithm is simulated for $30$ epochs, where each epoch  corresponds to $1000$ iterations.

{\bf MNIST $\{0,1,2,3,4\}$ to MNIST $\{5,6,7,8,9\}$.} To obtain the latent embeddings of the MNIST dataset, we first train a VAE with both the encoder and decoder having $3$ hidden layers with $256$ neurons and the size of latent vector being $16$ dimensional. We then use ICNNs $f$ and $g$ to learn the optimal transport between the embeddings of digits $\{0,1,2,3,4\}$ to that of $\{5,6,7,8,9\}$. For both these ICNNs we have $d=16, m=1024, L=3$, CELU activation, batch size = $128$, $K=16$, $\lambda=1$, $T=100,000$, Adam optimizer with $\beta_1=0.9$ and $\beta_2 =0.99$, learning rate decay by a factor of $0.5$ for every $4,000$ iterations.

{\bf Gaussian to MNIST.} To obtain the latent embeddings for the MNIST, we use the same pre-trained VAE models as above. Also we use the same hyperparameter settings as that of the ``{\bf MNIST $\{0,1,2,3,4\}$ to MNIST $\{5,6,7,8,9\}$}" experiment with the only change of batch size being $64$.

\section{Further discussion of related work}
\label{app:related-work-more}


The idea of solving the semi-dual optimization problem~\eqref{eq:dual_convex_form} is classically considered in~\cite{chartrand2009gradient}, where the authors derive a formula for the functional derivative of the objective function with respect to $f$ and propose to solve the optimization problem with the gradient descent method. Their approach is based on the discretization of the space and knowledge of the explicit form of the probability density functions, that is not applicable to real-world high dimensional problems. 

More recently,  the authors in~\cite{lei2017geometric,guo2019mode} propose to learn the function $f$ in a semi-discrete setting, where one of the marginals is assumed to be a discrete distribution supported on a  set of $N$ points $\{y_1,\ldots,y_N\} \subset \real^d $, and the other marginal is assumed to have a continuous density with compact convex support $\Omega \subset \real^d$. They show that the problem of learning the  function $f$ is similar to the variational formulation of the Alexandrov problem: constructing a convex polytope with prescribed face normals and volumes. Moreover, they show that, in the semi-distrete setting, the optimal $f$ is of the form $f(x) = \max_{1\leq i\leq 1}\{\langle x, y_i\rangle + b_i\}$ and simplify the problem of learning $f$ to the problem learning $N$ real numbers $b_i \in \real$. However, the objective function involves computing  polygonal partition of $\Omega$ into $N$ convex cells, induced by the function $f$, which is computationally challenging. Moreover, the learned optimal transport map $\nabla f$, transports the probability distribution from each convex cell to a single point $y_i$, which results in generalization issues. Additionally, the proposed approach is semi-discrete, and as a result, does not scale with the number of samples.

Statistical analysis of learning the optimal transport map through the semi-dual optimization problem~\eqref{eq:dual_convex_form} is studied in \cite{hutter2019minimax,rigollet2018uncoupled}, where the authors establish a minimax convergence rate with respect to number of samples for certain classes of regular probability distributions. They also propose a procedure that achieves the optimal convergence rate, that involves representing the function $f$ with span of wavelet basis functions up to a certain order, and also requiring the function $f$ to be convex. However, they do not provide a computational algorithm to implement the  procedure.

There are also other alternative approaches to approximate the optimal transport map that are not based on solving the semi-dual optimization problem~\eqref{eq:dual_convex_form}. In~\cite{leygonie2019adversarial}, the authors propose to approximate the optimal transport map, through an adversarial computational procedure, by considering the dual optimization problem~\eqref{eq:dual_form}, and replacing the constraint with a quadratic penalty term.
However, in contrast to the other regularization-based approaches such as~\cite{seguy2017large}, they consider a GAN architecture, and propose to take the generator, after the training is finished, as the optimal transport map. They also provide a theoretical justification for their proposal,  however the theoretical justification is valid in an ideal setting where the generator has infinite capacity, the discriminator is optimal at each update step,  and the cost is equal to the exact Wasserstein distance. These ideal conditions are far from being true in a practical setting.

Another approach, proposed in~\cite{xie2019scalable}, is to learn the optimal coupling from primal formulation~\eqref{eq:kantor_relax}, instead of solving the dual problem~\eqref{eq:dual_form}. The approach involves representing the coupling with two generators that map a Gaussian random variable  to $\real^d$, and two discriminators to ensure the coupling satisfies the marginal constraints. Although, the proposed approach is attractive when an optimal transport map does not exists, it is computationally expensive because it involves learning four deep neural networks. 
Finally, a procedure is recently proposed to approximate the optimal transport map that is optimal only on a subspace projection instead of the entire space~\cite{muzellec2019subspace}. This approach is  inspired by the sliced Wasserstein distance method to approximate the Wasserstein distance~\cite{rabin2011wasserstein,deshpande2018generative}. However, selection of the subspace to project on is a non-trivial task, and optimally selecting the projection is an optimization over the Grassmann manifold which is computationally challenging. 

In a recent work, \citet{korotin2019wasserstein} too model the convex conjugate function $f^*$ with an ICNN, denoted here by $g$, and a penalty term of the form $\|\nabla f(\nabla g(y)) -y\|^2$ is added to the semi-dual optimization~\eqref{eq:dual_convex_form}. The penalty term  serves to ensure that $\nabla g$ is inverse of $\nabla f$ and hence $g=f^*$. The additional penalty term makes the problem non-convex, even in the infinite capacity case, where the function representation is not restricted.

%
%
%

\end{document}